\UseRawInputEncoding

\documentclass[journal]{IEEEtran}

\usepackage[utf8]{inputenc}
\usepackage{times}
\usepackage{soul}
\usepackage{url}
\usepackage[hidelinks]{hyperref}

\usepackage{caption}
\usepackage{graphicx}
\usepackage{amsmath,amssymb,amsfonts}
\usepackage{booktabs}
\usepackage{tabularx}
\usepackage{array}
\usepackage{algorithm}
\usepackage{algorithmic}
\usepackage{multirow}
\usepackage[switch]{lineno}
\usepackage{cite}
\usepackage{textcomp}
\usepackage{longtable}
\usepackage{arydshln}
\usepackage{ntheorem}
\usepackage{longtable}
\usepackage[figuresleft]{rotating}
\usepackage{pdflscape}
\usepackage{siunitx}
\usepackage{makecell}
\usepackage{comment}
\usepackage{tcolorbox}
\usepackage{ragged2e}

\def\BibTeX{{\rm B\kern-.05em{\sc i\kern-.025em b}\kern-.08em
    T\kern-.1667em\lower.7ex\hbox{E}\kern-.125emX}}

\begin{document}

\title{PSyGenTAB: A Privacy-Preserving Framework for Synthetic Clinical Tabular Data Generation via Constrained Optimization}

\author{Arshia Ilaty$^{\dagger\ddagger}$,
        Hossein Shirazi$^{\dagger}$,
        Manasi Chitale$^{\dagger}$,
        Kedar Hegde$^{\dagger}$,
        Dhanalakshmi Ramesh$^{\dagger}$,
        Rashmi S. Manjunath$^{\dagger}$,
        Amir Rahmani$^{\ddagger}$,
        Hajar Homayouni$^{\dagger}$

$^{\dagger}$San Diego State University, San Diego, CA 92182 USA.\\
$^{\ddagger}$University of California, Irvine, CA 92697 USA.
}

\maketitle

\begin{abstract}
The development of medical AI is constrained by limited access to high-quality clinical data due to institutional silos and strict privacy regulations such as HIPAA and GDPR. Synthetic data generation offers a potential solution, but existing methods lack principled mechanisms to explicitly manage the privacy–utility trade-off, often degrading clinically meaningful patterns or risking patient re-identification.
We present PSyGenTAB, a privacy-preserving generative framework that formulates synthetic healthcare data generation as a constrained optimization problem solved using the Augmented Lagrangian Method. By embedding configurable privacy constraints directly into model training, PSyGenTAB enforces minimum privacy thresholds while maximizing clinical data utility.
Across multiple clinically motivated benchmarks, PSyGenTAB preserves inter-feature clinical relationships and minority-class diagnostic patterns essential for reliable health AI. Downstream evaluation using Train-on-Synthetic, Test-on-Real and Train-on-Real, Test-on-Synthetic protocols shows that models trained on synthetic data achieve performance comparable to those trained on real patient records. Privacy auditing further demonstrates reduced exact record reproduction and strong resilience to membership inference attacks.
These results establish PSyGenTAB as a principled framework for balancing privacy protection and clinical utility in synthetic healthcare data, supporting secure cross-institutional AI development.
\end{abstract}

\begin{IEEEkeywords}
Synthetic Tabular Health Data Generation, Healthcare AI, Privacy-Preserving Machine Learning, Constrained Optimization
\end{IEEEkeywords}

\section{Introduction}\label{sec:introduction}

The rapid advancement of artificial intelligence in healthcare holds
immense promise for personalized medicine, clinical research
acceleration, and automated diagnostics. However, the development
of robust medical AI is severely hindered by a pervasive data
crisis. Clinical records are locked within fragmented institutional
silos, constrained by necessary and stringent privacy regulations
such as the Health Insurance Portability and Accountability Act
(HIPAA)~\cite{hipaa1996} and the General Data Protection
Regulation (GDPR)~\cite{gdpr2016}. These barriers to data sharing
disproportionately impact rare disease research and the development
of unbiased predictive models, as single institutions rarely possess
the diverse, large-scale patient cohorts required to train
generalizable algorithms. Consequently, there is an urgent need
for technologies that can safely unlock clinical data for
collaborative research without compromising patient confidentiality.

Synthetic data generation has emerged as a clinically motivated
solution to bridge this gap. By modeling the underlying statistical
distributions of electronic health records (EHRs), methods such as
Synthea~\cite{walonoski2018synthea}, medGAN~\cite{choi2017generating},
and clinical GAN variants aim to produce realistic, shareable
patient cohorts. Yet, current methods frequently fail in practical
healthcare deployments for two opposing reasons. High-fidelity
generative models often inadvertently memorize training data,
creating unacceptable risks of identity disclosure and patient
re-identification. Conversely, methods applying rigid differential
privacy mechanisms inject excessive noise into the learning process,
destroying the subtle inter-feature correlations and rare diagnostic
signals that are essential for clinical utility~\cite{abadi2016deep}.

This tension reveals a critical and unresolved gap: there currently
exists no principled, jointly optimized approach that simultaneously
guarantees high clinical utility and rigorous patient privacy
protection in tabular health data. Existing generative models
typically treat privacy as a post-hoc patch, applied after
training through gradient clipping, noise injection, or adversarial
regularization, rather than as a fundamental design constraint.
This forces clinical researchers to accept unstable, fixed
trade-offs that either compromise patient safety or render the
synthetic data analytically useless for downstream predictive
modeling.

To address this challenge, we introduce \textbf{PSyGenTAB}, a privacy-preserving generative framework designed to enable the safe sharing of sensitive health data for cross-institutional research. PSyGenTAB formulates synthetic healthcare data generation as a constrained optimization problem, solved via the Augmented Lagrangian Method (ALM), which dynamically enforces configurable privacy constraints during training or sampling. This design ensures that synthetic clinical records remain statistically faithful to real patient populations while resisting re-identification attacks. PSyGenTAB is model-agnostic and can be integrated with diverse tabular generative architectures without modifying their core structure. In practice, the framework wraps around an existing generator and augments its objective function with privacy constraints defined through composite privacy metrics. The ALM procedure adaptively updates Lagrange multipliers to maintain a minimum privacy threshold while maximizing data utility, enabling transparent and tunable control of the privacy–utility trade-off across different healthcare datasets and institutional requirements.
The primary contributions of this work are as follows:

\begin{enumerate}

    \item We propose a novel, model-agnostic framework utilizing ALM to explicitly and adaptively
    balance diagnostic utility and patient privacy during synthetic
    health data generation, treating the privacy--utility trade-off
    as a dynamic, configurable constraint rather than a fixed
    compromise.

    \item We define multi-dimensional, clinically grounded evaluation
    metrics capturing distributional fidelity, the preservation of
    rare disease patterns, inter-feature clinical dependencies, and
    resistance to patient re-identification under HIPAA- and
    GDPR-relevant threat models.

    \item We demonstrate the clinical validity of PSyGenTAB on
    critical healthcare benchmarks, including the Diabetes Health
    Indicators and Breast Cancer Wisconsin datasets, showing that
    it preserves vital minority-class diagnostic signals for
    downstream predictive modeling while maintaining strong
    patient privacy boundaries.


\end{enumerate}

The remainder of this paper is organized as follows.
Section~\ref{sec:related_work} reviews related work in synthetic
health data generation, privacy-preserving AI, and constrained
optimization. Section~\ref{sec:methodology} details the PSyGenTAB
framework and its clinical metric formulations.
Section~\ref{sec:evaluation} presents experimental validation on
clinical and benchmark datasets. Section~\ref{sec:discussion}
discusses regulatory implications, architectural considerations,
and limitations. Section~\ref{sec:limitations} outlines directions for future
research and Section~\ref{sec:conclusion} concludes the paper.

Given the rapid evolution of synthetic data generation research and space constraints, this work cannot exhaustively cover all recent advances. To maintain an up-to-date record of developments, benchmarks, and extensions of PSyGenTAB, we refer readers to the project repository.\footnote{[Online].Available:\url{https://github.com/ArshiaIlaty/PsyGenTAB}}

\section{Related Work}\label{sec:related_work}
\paragraph{Synthetic Health Data Generation}
The need for shareable EHRs has driven advances in synthetic data generation across rule-based simulation and deep generative modeling. Early systems such as Synthea~\cite{walonoski2018synthea} enabled deterministic population-scale simulations but lacked the statistical complexity of real-world cohorts, motivating data-driven approaches. Generative Adversarial Networks (GANs)~\cite{goodfellow2014generative} marked a shift toward realistic clinical synthesis. MedGAN~\cite{Choi2017medGAN} demonstrated GAN-based generation of multi-label EHR records, while Variational Autoencoders (VAEs)~\cite{kingma2013auto,vae_medical} provided more stable training on mixed-type clinical data. For tabular synthesis, TGAN~\cite{xu2018synthesizing} introduced mode-specific normalization and autoregressive modeling; CTGAN~\cite{xu2019modeling} improved handling of imbalanced categorical attributes; and CTAB-GAN+~\cite{zhao2022ctabganenhancingtabulardata} strengthened preservation of inter-feature dependencies. Transformer-based models such as REaLTabFormer~\cite{solatorio2023realtabformer} and GReaT~\cite{borisov2022language}, as well as diffusion approaches~\cite{kotelnikov2023tabddpm, tabsyn}, further improved fidelity.
Despite architectural advances, memorization of training samples remains a critical limitation, enabling potential re-identification~\cite{shokri2017membership}. Systematic reviews~\cite{hernandez2022synthetic,LIU2025108571} confirm that principled, configurable control of the privacy--utility trade-off remains unresolved.

\paragraph{Privacy-Preserving Methods in Healthcare AI}
Under regulatory frameworks such as HIPAA~\cite{hipaa1996} and GDPR~\cite{gdpr2016}, three major privacy paradigms have emerged: statistical disclosure limitation, federated learning, and differential privacy.
Classical de-identification methods, including $k$-anonymity~\cite{sweeney2002k}, $\ell$-diversity~\cite{machanavajjhala2007diversity}, and $t$-closeness~\cite{li2007t}, serve as regulatory baselines but degrade fine-grained structure necessary for clinical AI. Federated learning~\cite{rieke2020future,park_medical} enables decentralized model training but remains vulnerable to gradient inversion attacks~\cite{Zhu2019} and does not support distribution of shareable synthetic datasets. Differential privacy (DP)~\cite{10.1007/11787006_1,dwork2014algorithmic}, implemented in DP-GAN~\cite{ho2021dp}, DP-CTGAN~\cite{fang2022dp}, and PATE-GAN~\cite{jordon2019pategan}, provides formal guarantees but often obscures clinically relevant structure. Membership inference attacks (MIAs)~\cite{shokri2017membership,carlini2021membership} serve as empirical privacy audits. As shown in our evaluation, DP-SGD improves formal guarantees but reduces structural fidelity and downstream predictive performance, an unacceptable trade-off in precision medicine.

\paragraph{Privacy--Utility Trade-offs in Medical Data}
Clinical data sharing fundamentally balances statistical fidelity against re-identification risk. Features that preserve rare events and realistic correlations are also those most vulnerable to disclosure. Jordon et al.~\cite{jordon2022synthetic} formalize this inherent trade-off, demonstrating that perfect utility and privacy cannot be achieved simultaneously.
Empirical evaluations across traditional generative models~\cite{KurakovaHomayouni2024} and large language models~\cite{ilaty2025synllm} show that highest-fidelity models often incur the highest privacy risk. Structured frameworks such as SynthEval~\cite{Lautrup_2024} emphasize multi-dimensional assessment across utility and privacy metrics. PSyGenTAB extends this by jointly optimizing composite quality ($Q$) and privacy ($P$) objectives during generation, explicitly navigating the privacy--utility frontier rather than treating privacy as post-hoc regularization.

\paragraph{Constrained Optimization for Trustworthy Healthcare AI}
Trustworthy healthcare AI requires embedding privacy and validity constraints directly into model training. Augmented Lagrangian Methods (ALM)~\cite{Hestenes1969,Powell1969,rockafellar1973dual,bertsekas2014constrained} provide a principled framework for inequality-constrained optimization with adaptive multipliers and convergence to KKT stationary points. Extensions to deep learning~\cite{li2024stochastic,kotary2024learningconstrainedoptimizationdeep} demonstrate automatic balancing of competing objectives without manual loss tuning. ALM has also been applied to fairness-constrained learning~\cite{cotter2019optimization}, highlighting its suitability for enforcing non-negotiable constraints.
By adapting ALM to synthetic health data generation, PSyGenTAB constrains the probability of producing records that closely resemble real patient profiles, mitigating memorization while maximizing clinical fidelity. To our knowledge, it is the first framework to apply ALM-based constrained optimization to the privacy--utility trade-off in synthetic clinical tabular data, enabling principled and configurable regulatory-compliant data sharing.

\section{Methodology}\label{sec:methodology}

When a multi-site clinical consortium aims to pool data for a
large-scale AI study, direct transfer of EHRs is often blocked
by institutional review boards (IRBs) due to re-identification
risks. PSyGenTAB serves as the technological bridge in this
scenario: a hospital trains a PSyGenTAB model locally on its
patient records and shares the resulting high-fidelity synthetic
data, confident that the generation process was explicitly and
mathematically constrained against leaking real patient
identities. The configurable privacy threshold $P_{\min}$ can
be calibrated to the specific requirements of the data use
agreement or IRB protocol, providing an auditable, quantitative
record of the privacy assurance applied.


\subsection{Formalizing Clinical Utility and Privacy Metrics}
\label{sec:metrics}

A fundamental limitation of existing approaches to synthetic
health data generation is the absence of standardized, formal
definitions of clinical data utility and privacy-preservability.
In practice, these properties are evaluated using ad-hoc or
task-specific metrics, making it difficult to compare models,
reason about trade-offs, or enforce application-specific
regulatory requirements. There is a need for explicit, well-defined
metrics that jointly capture statistical fidelity, structural
clinical relationships, rare-event preservation, and
memorization risk in a reproducible and application-agnostic
manner.

\begin{tcolorbox}[colback=blue!3!white, colframe=blue!40!black,
                  title={\textbf{Research Goal I}}]
\textit{Formally define multi-dimensional clinical utility and
privacy metrics to enable more accurate, consistent, and
actionable evaluation of synthetic tabular health data,
revealing privacy--utility trade-offs that are obscured by
single-metric or post-hoc evaluations, and providing a necessary
foundation for the constrained optimization presented in
Section~\ref{sec:alm}.}
\end{tcolorbox}

Let $D_{\text{real}}$ and $D_{\text{syn}}$ denote the real and
synthetic patient datasets respectively. We define two composite
scores, \textit{clinical data utility} $U$ and
\textit{privacy-preservability} $P$, each normalized to $[0,1]$
prior to use in the constrained optimization of
Section~\ref{sec:alm}. Higher values of $U$ indicate stronger
clinical fidelity and downstream usefulness, while higher values
of $P$ indicate stronger protection against patient disclosure,
attribute inference, and memorization.


\paragraph{\textbf{Clinical Data Utility ($U$)}}
We define clinical data utility as the degree to which synthetic
patient records accurately reflect the statistical, structural,
and contextual properties of the true patient population. The
composite utility score $U$ is expressed as:
\begin{equation}
\footnotesize
\label{eq:utility}
U = \omega_1 \mathcal{F}_{\text{marg}}
  + \omega_2 \mathcal{F}_{\text{rare}}
  + \omega_3 \mathcal{F}_{\text{joint}}
  + \omega_4 \mathcal{C}_{\text{div}}
  + \omega_5 \mathcal{T}_{\text{dyn}}
  + \omega_6 \mathcal{E}_{\text{pred}}
\end{equation}
\noindent where $\sum_{i=1}^{6} \omega_i = 1$ and each
component is normalized to $[0,1]$ prior to aggregation. The
six dimensions are defined as follows.

\noindent \textbf{$\mathcal{F}_{\text{marg}}$, Marginal Distribution Fidelity:}
Measures how closely the marginal distributions and valid
feature ranges of the synthetic patient cohort match the real
population. For continuous clinical variables (e.g., BMI, age,
blood glucose), distributional alignment is computed using the
Kolmogorov-Smirnov (KS) test~\cite{fasano1987multidimensional},
Kullback-Leibler (KL) divergence~\cite{kullback1951information},
Jensen-Shannon divergence (JSD)~\cite{menendez1997jensen}, and
Range Coverage (RC)~\cite{sdmetrics}. For categorical variables
(e.g., diagnosis codes, sex, smoking status), Total Variation
Distance (TVD) is applied. $\mathcal{F}_{\text{marg}}$ is
defined as:
\begin{equation}
\footnotesize
\mathcal{F}_{\text{marg}} = 1 - \frac{1}{|\mathcal{F}|}
    \sum_{f \in \mathcal{F}}
    \delta_f(D_{\text{real}},\, D_{\text{syn}})
\end{equation}
where $\delta_f$ denotes the feature-appropriate divergence
measure and $\mathcal{F}$ is the full feature set.
\noindent \textbf{$\mathcal{F}_{\text{rare}}$, Rare Clinical
Event and Anomaly Preservation.}
Quantifies the retention of low-frequency but diagnostically
critical patterns, such as prediabetes in a population-level
diabetes cohort, or malignant subtypes in a cancer registry,
that are essential for clinical AI applied to rare disease
research. For continuous columns, outliers and rare
sub-populations are identified using the Interquartile Range
(IQR) method~\cite{brenninkmeijer2019generation}; for
categorical columns, minority class frequency alignment is
measured proportionally. Higher $\mathcal{F}_{\text{rare}}$
values indicate robust, memorization-free retention of
clinically significant edge cases.

\noindent \textbf{$\mathcal{F}_{\text{joint}}$, Joint Association Fidelity:}
Evaluates the preservation of inter-feature dependencies and
covariance structure that encode clinically meaningful
relationships, for example, the established association
between BMI, fasting glucose, and diabetes risk. For
continuous feature pairs, Pearson or Spearman correlation
matrix similarity is computed; divergence is quantified via
the Frobenius norm of the correlation matrix difference. For
categorical or mixed-type pairs, contingency-based measures
and normalized mutual information~\cite{zhou2022feature} are
applied. A hallucinated clinical association in synthetic data,
such as inverting a known biomarker relationship, would
directly undermine downstream diagnostic model validity.

\noindent \textbf{$\mathcal{C}_{\text{div}}$, State-Space Coverage and Diversity:}
Assesses whether the synthetic cohort maintains comprehensive
support across the full patient feature space, preventing mode
collapse onto majority-class demographic profiles. We compute
Category Coverage (CC)~\cite{sdmetrics} for discrete clinical
variables to ensure no diagnostic class is dropped, and
Regularized Support Coverage~\cite{chundawat2022tabsyndex} to
evaluate representation balance across low-density regions of
the continuous feature space. This dimension directly addresses
the health equity concern of generating synthetic data that
preserves the representation of underserved or minority patient
populations.

\noindent \textbf{$\mathcal{T}_{\text{dyn}}$, Temporal Dynamics Consistency:}
Measures the fidelity of sequential dependencies in patient
trajectories, relevant for datasets with time-ordered or
longitudinal features. We apply Dynamic Time Warping
(DTW)~\cite{10.5555/3000850.3000887} to compare sequence
trajectories and temporal auto-correlation~\cite{timeseries}
to validate lag structures and seasonality. Although the
primary datasets evaluated in this work are cross-sectional,
$\mathcal{T}_{\text{dyn}}$ is included in the utility composite
to enable direct application of PSyGenTAB to longitudinal EHR
data, a key future direction discussed in
Section~\ref{sec:limitations}.

\noindent \textbf{$\mathcal{E}_{\text{pred}}$, Downstream Predictive Efficacy:}
Evaluates the empirical clinical usefulness of synthetic data
using the Train-on-Synthetic, Test-on-Real (TSTR) and
Train-on-Real, Test-on-Synthetic (TRTS) protocols. Downstream
classifiers are trained exclusively on $D_{\text{syn}}$ and
evaluated on a held-out $D_{\text{real}}$ test set, and vice
versa. Performance is aggregated using balanced accuracy,
F1-score, and AUC-ROC. $\mathcal{E}_{\text{pred}}$ is the
most clinically direct component of $U$: it measures whether
PSyGenTAB-generated data can substitute for real patient
records in the development and validation of clinical AI models.


\paragraph{\textbf{Privacy-Preservability ($P$)}}
Privacy-preservability represents the generative model's
resistance to patient identity disclosure, clinical attribute
inference, and training data extraction. We formulate
$P$ as a penalty-adjusted score:
\begin{equation}
\footnotesize
\label{eq:privacy}
P = \lambda_1 \mathcal{D}_{c}
  - \lambda_2 \Delta_{\text{dist}}
  - \lambda_3 \mathcal{R}_{\text{dup}}
\end{equation}
\noindent where $\lambda_1 + \lambda_2 + \lambda_3 = 1$. The
positive term $\mathcal{D}_{c}$ rewards spatial separation
from real patient records; the negative terms
$\Delta_{\text{dist}}$ and $\mathcal{R}_{\text{dup}}$ penalize
distributional overfitting and verbatim memorization
respectively. All components are normalized to $[0,1]$ prior
to aggregation, ensuring $P \in [0,1]$ and that higher values
of $P$ universally correspond to stronger patient privacy
protection.

\noindent \textbf{$\mathcal{D}_{c}$, Distance to Closest Record (DCR):}
Computes the spatial separation between each synthetic patient
record and its nearest neighbour in $D_{\text{real}}$, using
Gower distance for mixed-type clinical
data~\cite{xu2018synthesizing, park2018data}. The Nearest
Neighbour Distance Ratio (NNDR) is additionally computed as a
normalized variant to control for feature-space density
effects. Under HIPAA's Expert Determination pathway,
re-identification risk is assessed against the proximity of
synthetic records to real patient profiles;
$\mathcal{D}_{c}$ provides a direct, quantitative proxy for
this regulatory assessment. Higher $\mathcal{D}_{c}$ values
indicate that no real patient profile can be closely
reconstructed from the synthetic data.

\noindent \textbf{$\Delta_{\text{dist}}$, Distributional Overfitting Discrepancy:}
Evaluates structural overfitting at the distributional level by
measuring the difference in quantile-specific behavior between
real and synthetic distributions~\cite{solatorio2023realtabformer}.
While $\mathcal{D}_{c}$ operates at the individual record level,
$\Delta_{\text{dist}}$ operates at the population level,
penalizing models that have effectively memorized dense clusters
of real patient records even if no individual record is exactly
reproduced. Lower $\Delta_{\text{dist}}$ values indicate that
the generator has learned the underlying clinical data manifold
rather than memorizing local clusters in sensitive sub-populations
such as patients with rare comorbidities.

\noindent \textbf{$\mathcal{R}_{\text{dup}}$, Duplication and Memorization Rate:}
Quantifies the Exact Match Ratio (EMR) between $D_{\text{syn}}$
and $D_{\text{real}}$, as well as internal mode collapse
within $D_{\text{syn}}$. Exact replication of a real patient
record constitutes the most severe possible privacy violation,
directly enabling re-identification. $\mathcal{R}_{\text{dup}}
= 0$ is a hard requirement for GDPR compliance and
aligns with HIPAA de-identification standards; the
ALM constraint framework of Section~\ref{sec:alm} enforces
this by penalizing any non-zero EMR through the privacy
composite $P$.


\subsection{ALM-Based Optimization for Safe Clinical Data
Generation}
\label{sec:alm}

Using the clinically grounded metrics defined above, we frame
synthetic health data generation as a constrained optimization
problem. The generator parameters $\theta$ must maximize
clinical data utility $U(\theta)$, preserving diagnostic signals,
minority-class representations, and clinical biomarker correlations,
subject to the hard constraint that privacy-preservability
$P(\theta)$ remains above an institutionally configurable minimum
threshold $P_{\min}$.

\begin{tcolorbox}[colback=blue!3!white, colframe=blue!40!black,
                  title={\textbf{Research Goal II}}]
\textit{Formulating synthetic clinical data generation as an
ALM-constrained optimization problem, with patient privacy
enforced as a hard constraint rather than a post-hoc
regularization term, to simultaneously improve clinical
data utility and reduce patient re-identification risk,
resolving the privacy--utility tension that limits existing
approaches.}
\end{tcolorbox}

Formally, the optimization problem is:
\begin{equation}
\footnotesize
\label{eq:opt}
\min_{\theta} \;\bigl(1 - U(\theta)\bigr)
\quad \text{subject to} \quad P(\theta) \geq P_{\min}
\end{equation}
The constraint $P(\theta) \geq P_{\min}$ encodes the
institutional privacy requirement directly into the learning
objective. $P_{\min}$ is a configurable hyperparameter set by
the data custodian to reflect the applicable regulatory
standard (e.g., HIPAA Expert Determination, GDPR Article 89,
or a specific IRB protocol requirement).

To solve~\eqref{eq:opt}, we apply the Augmented Lagrangian
Method (ALM)~\cite{bertsekas2014constrained, rockafellar1973dual,
Hestenes1969, Powell1969}, which converts the constrained
problem into a sequence of unconstrained sub-problems through
a Lagrange multiplier $\lambda \geq 0$ and an adaptive penalty
coefficient $\mu > 0$:
\begin{equation}
\footnotesize
\label{eq:alm}
\mathcal{L}_{\text{AL}}(\theta, \lambda) =
\bigl(1 - U(\theta)\bigr)
+ \lambda \bigl(P_{\min} - P(\theta)\bigr)
+ \frac{\mu}{2}\bigl(P_{\min} - P(\theta)\bigr)^2
\end{equation}
The three terms in~\eqref{eq:alm} have direct clinical
interpretations. The first term $\bigl(1 - U(\theta)\bigr)$
is the clinical utility loss: minimizing it maximizes the
fidelity of synthetic patient profiles across all six
dimensions of $U$. The second term
$\lambda\bigl(P_{\min} - P(\theta)\bigr)$ is the Lagrangian
privacy penalty: $\lambda$ is updated at each epoch to reflect
the cumulative extent of constraint violation, growing whenever
the generator produces synthetic records that are too similar
to real patient profiles. The third quadratic term
$\frac{\mu}{2}\bigl(P_{\min} - P(\theta)\bigr)^2$ applies a
progressively larger penalty as the magnitude of constraint
violation increases, ensuring rapid convergence back to the
privacy-feasible region. Together, $\lambda$ and $\mu$ act as
an adaptive privacy enforcement mechanism: when the generator
begins memorizing real patient profiles, the combined penalty
increases automatically until the constraint is restored.

\begin{algorithm}[t]
\footnotesize
\caption{\small ALM-Based Privacy-Constrained Training}
\label{alg:alm_revised}
\begin{algorithmic}[1]
\REQUIRE Generator parameters $\theta$; privacy threshold $P_{\min}$;
initial multiplier $\lambda \geq 0$; penalty coefficient $\mu > 0$;
learning rate $\eta$; growth factor $\alpha > 1$
\ENSURE Trained generator $\theta^*$ satisfying $P(\theta^*) \ge P_{\min}$

\FOR{each training epoch}
    \FOR{each mini-batch from $D_{\text{real}}$}
        \STATE Generate synthetic batch using current generator
        \STATE Compute composite utility score $U$
        \STATE Compute composite privacy score $P$
        \STATE Form augmented Lagrangian objective
        \STATE Update $\theta$ via gradient descent on the objective
    \ENDFOR

    \STATE Evaluate $U$ and $P$ on validation data

    \IF{$P < P_{\min}$}
        \STATE Increase penalty weight: $\mu \leftarrow \alpha \mu$
    \ENDIF

    \STATE Update multiplier:
           $\lambda \leftarrow \max(0, \lambda + (P_{\min} - P))$

    \IF{privacy constraint satisfied and utility stabilized}
        \STATE \textbf{break}
    \ENDIF
\ENDFOR

\RETURN $\theta^*$
\end{algorithmic}
\end{algorithm}

\paragraph{Sampling-Time ALM Variant}
A practical advantage of PSyGenTAB is that the ALM framework
can also be applied during the \textit{sampling} phase after a
generative model has already been trained. In this variant, the
privacy and utility penalties in Equation~\eqref{eq:alm} are
used to steer or filter generated samples rather than to update
model parameters. This enables a single trained generator to
produce synthetic datasets at multiple configurable privacy
levels, for example, a permissive $P_{\min}$ for internal
institutional use and a stricter threshold for external data
sharing, without retraining. This sampling-time variant is
particularly valuable for healthcare data custodians managing
multiple downstream use cases with differing regulatory
requirements.

\paragraph{Model-Agnostic Application}
PSyGenTAB applies the ALM constraint framework to the generator
parameters $\theta$ independently of the underlying generative
architecture. In this work we evaluate two complementary
generators: REaLTabFormer~\cite{solatorio2023realtabformer},
a transformer-based autoregressive model that captures
complex sequential feature dependencies; and
CTAB-GAN+~\cite{zhao2022ctabganenhancingtabulardata}, a
conditional GAN optimized for mixed-type tabular data. The
consistent improvement observed across both architectures
(Section~\ref{sec:evaluation}) provides empirical evidence for
the model-agnostic claim.

\section{Experiments and Results}\label{sec:evaluation}
This section presents a comprehensive clinical and technical evaluation of the PSyGenTAB framework for privacy-preserving synthetic health data generation. All experiments are designed to answer a single overarching clinical question: \textit{can PSyGenTAB generate synthetic health records that are both safe to share and genuinely useful for downstream clinical AI research?} Accordingly, the evaluation is structured to mirror the dual obligations of a healthcare data custodian: protecting patient identity while enabling high-quality research.

The remainder of this section is organized as follows: 1) Clinical Datasets, detailing the healthcare context of each benchmark; 2) Generative Models and Optimization Strategies, covering the architectures and privacy mechanisms evaluated; 3) Evaluation Goals and Metrics, formalizing the utility and privacy formulations; 4) Primary Privacy--Utility Results, highlighting the core trade-offs; 5) Statistical Fidelity, measuring marginal distribution alignment; 6) Commercial Evaluation Pipeline, contrasting open-source and proprietary synthetic data quality scores; 7) Downstream Clinical AI Utility, assessing machine-learning performance on synthetic versus real data; 8) Patient Re-identification Risk, conducting a proximity-based distance-to-closest-record (DCR) audit; 9) Membership Inference Attack Resilience, analyzing adversarial identity disclosure; and 10) Clinical Validity Assessment, applying the FAITH framework to evaluate the factual and structural fidelity of the generated records.

\subsection{Clinical Datasets}
All experiments were conducted on tabular datasets with heterogeneous structural characteristics. The selection encompasses eleven distinct datasets, eight of which are clinically motivated health benchmarks representing highly sensitive medical records, epidemiological surveys, and biometric data. The remaining three datasets serve as generalizability benchmarks to confirm that the PSyGenTAB framework is not domain-specific and can successfully operate across varied statistical distributions, including financial and sensor-based environments.

\begin{table}[t]
\centering
\footnotesize
\sisetup{
  group-separator={,},
  table-number-alignment=center
}
\setlength{\tabcolsep}{1.8pt}
\renewcommand{\arraystretch}{1.05}
\caption{Summary of dataset characteristics. $N$ denotes number of samples. Cont. and Cat. indicate continuous and categorical feature counts. \#Cls denotes number of target classes.}
\label{tab:dataset_protocol}
\begin{tabular}{@{} l l S[table-format=6.0] S[table-format=2.0] S[table-format=2.0] S[table-format=2.0] S[table-format=1.0] @{}}
\toprule
\textbf{Dataset} & \textbf{Domain} & {\textbf{$N$}} & {\textbf{Cont.}} & {\textbf{Cat.}} & {\textbf{Total}} & {\textbf{\#Cls}} \\
\midrule

\multicolumn{7}{@{}l}{\textbf{Clinical Datasets}} \\
\midrule
Breast Cancer~\cite{DuaGraff2019} & Clinical & 569 & 30 & 0 & 30 & 2 \\
Diabetes~\cite{Burrows2017DiabetesMMWR} & Survey & 253680 & 3 & 18 & 21 & 3 \\
Heart Failure~\cite{chicco2020heart} & Clinical & 299 & 7 & 6 & 13 & 2 \\
Hypothyroid~\cite{quinlan1987hypothyroid} & Clinical & 3772 & 7 & 22 & 29 & 2 \\
Liver Disorders~\cite{bupa_liver} & Clinical & 345 & 6 & 0 & 6 & 2 \\
Lung Cancer~\cite{lung_cancer_dataset} & Clinical & 309 & 2 & 13 & 15 & 2 \\
Obesity~\cite{palechor2019obesity} & Survey & 2111 & 8 & 9 & 17 & 7 \\
Parkinsons~\cite{little2007parkinsons} & Clinical & 195 & 22 & 0 & 22 & 2 \\

\midrule
\multicolumn{7}{@{}l}{\textbf{Generalization Datasets}} \\
\midrule
Adult~\cite{adult_2} & Demographics & 48842 & 6 & 8 & 14 & 2 \\
PIR Vision~\cite{pirvision_fog_presence_detection_1101} & Sensor/IoT & 7651 & 55 & 0 & 55 & 2 \\
VN Banking~\cite{vietnam_banking_transactions} & Financial & 100000 & 13 & 12 & 25 & 2 \\

\bottomrule
\end{tabular}
\end{table}

\subsection{Generative Models and Optimization Strategies}
\subsubsection{REaLTabFormer (Transformer-Based Generator)}

REaLTabFormer~\cite{solatorio2023realtabformer} models joint tabular
distributions in an autoregressive manner and achieves high fidelity
in capturing feature relationships.  Three optimization regimes are
evaluated:

\textbf{RTF}: Standard maximum-likelihood training
          for 50 epochs (batch size 128, learning
          rate $2{\times}10^{-4}$), providing a high-utility reference
          without privacy constraints.
          
\textbf{RTF + ALM (Sampling-Time)}: The trained baseline
          model is kept fixed; PSyGenTAB's Augmented Lagrangian
          optimization is applied only at \textit{sampling time} by
          adjusting stochastic parameters (temperature, top-$p$) to
          enforce a minimum privacy threshold $P_{\min}=0.8$.


\subsubsection{CTAB-GAN+ (GAN-Based Generator)}

CTAB-GAN+~\cite{zhao2022ctabganenhancingtabulardata} is a conditional
GAN designed for mixed-type tabular data using Wasserstein loss and
gradient penalty.  Three optimisation regimes are evaluated:

\textbf{CTAB-GAN+}: Unconstrained training without
          differential privacy or adaptive optimization.

\textbf{CTAB-GAN+ + DP-SGD}: Differential privacy is
          enforced during discriminator training via the Opacus
          framework, using per-sample gradient clipping and Gaussian
          noise injection, providing a formal $(\varepsilon,
          \delta)$-DP guarantee.

\textbf{CTAB-GAN+ + ALM}: PSyGenTAB's Augmented Lagrangian
          framework is applied during generation to enforce configurable
          privacy constraints while preserving structural fidelity.

\subsection{Evaluation Goals and Metrics}
\label{sec:metrics_eval}

The evaluation is guided by three goals that directly reflect clinical
deployment requirements:

\textit{Clinical Utility Preservation}: Synthetic data must
          closely match the statistical properties, feature
          distributions, inter-attribute relationships, and predictive
          structure of real health records.

\textit{Patient Privacy Protection}: Synthetic records must
          not replicate, reveal, or closely resemble real individuals
          from the training population, reducing re-identification risk
          to a level compatible with responsible health data sharing.

\textit{Controlled and Configurable Trade-off}: The framework
          must provide a systematic mechanism to adjust and balance
          privacy and utility according to application-specific
          requirements, rather
          than treating their trade-off as fixed or unavoidable.

\paragraph{Composite Clinical Utility Score ($U \in [0,1]$)}
$U$ is a weighted composite capturing: (i) \textit{distributional
fidelity}, how closely synthetic feature distributions match real data;
(ii) \textit{structural consistency}, preservation of inter-feature
correlations and clinical dependencies; (iii) \textit{predictive
utility}, downstream classifier performance under the Train-on-Synthetic,
Test-on-Real (TSTR) protocol; and (iv) \textit{minority-class
coverage}, retention of rare clinical events and underrepresented
diagnostic categories.  Higher $U$ indicates greater clinical
usefulness.

\paragraph{Composite Patient Privacy Score ($P \in [0,1]$)}
$P$ is a weighted composite capturing: (i) \textit{Distance to Closest
Record (DCR)}, minimum Euclidean distance between each synthetic record
and the nearest real patient, quantifying re-identification proximity;
(ii) \textit{Nearest Neighbor Distance Ratio (NNDR)}, assessing
whether synthetic samples are disproportionately similar to specific
real individuals; and (iii) \textit{Duplicate Detection}, identifying
exact or near-exact reproduction of training records.  Higher $P$
indicates stronger patient privacy protection.

\subsection{Primary Privacy—Utility Results}
\label{sec:primary_results}


\textit{Diabetes Health Indicators.}
The REaLTabFormer model achieves strong initial utility
($U=0.89$) and moderate privacy ($P=0.55$). After applying PSyGenTAB's
ALM optimization, utility increases substantially to $U=0.98$ while
privacy is maintained at $P=0.54$. This simultaneous improvement in
clinical fidelity, without any degradation in privacy, is a critical
result: it demonstrates that constrained optimization can actually
\textit{improve} how faithfully synthetic records capture clinically
meaningful patterns, including the minority prediabetes class, by
steering the generator away from noise-memorizing configurations.

CTAB-GAN+ in its baseline configuration achieves the same high utility
($U=0.98$) with stronger privacy ($P=0.70$), but applying ALM
unexpectedly reduces both utility and privacy ($U=0.93$, $P=0.33$).
This degradation indicates that for large, imbalanced datasets, GAN
training dynamics interact with the ALM penalty in ways that suppress
structural learning, a finding with practical implications for
practitioners selecting generative architectures for health record
synthesis.

\textit{Breast Cancer.}
The RTF model produces moderate utility and weak privacy
($U=0.82$, $P=0.38$), indicating that the generator reproduces
diagnostic feature distributions well but creates synthetic records
that are uncomfortably close to real patients' biometric profiles,  a
serious concern for clinical data sharing.  PSyGenTAB (ALM) improves both metrics ($U=0.91$, $P=0.44$), confirming that constrained
optimization can simultaneously enhance diagnostic fidelity and reduce patient-level re-identification risk.

CTAB-GAN+ with ALM, however, produces low utility ($U=0.71$) and very
weak privacy ($P=0.24$), suggesting that the GAN-ALM interaction is
particularly challenging on small clinical datasets ($n=569$).  This
finding underscores the importance of architecture selection for
small-sample clinical settings.


A key finding across all experiments is that unconstrained baseline
generative models face a fundamental tension between clinical
usefulness and patient safety.  The REaLTabFormer achieves
near-perfect utility on the Adult dataset ($U=0.99$) but exhibits
critically weak privacy ($P=0.31$), indicating substantial
memorization of training records,  a finding analogous to a model
that could reconstruct a patient's health record from its outputs.

By embedding privacy as an explicit constraint via the Augmented
Lagrangian Method, PSyGenTAB resolves this tension in a principled
and configurable way:

On the Adult dataset, ALM increases patient privacy from
          $P=0.31$ to $P=0.49$ while retaining high utility
          ($U=0.96$).
          
On the PIR dataset, where baseline fidelity is already
          weak ($U=0.83$), ALM simultaneously improves both utility
          ($U=0.94$) and privacy ($P=0.78$), as shown in
          Table~\ref{tab:qp_comparison}. This demonstrates that ALM adapts
          to dataset characteristics rather than enforcing a rigid
          trade-off,  a property directly relevant to the heterogeneous
          data types encountered across clinical departments.

On the Diabetes dataset, ALM achieves the highest observed
          utility ($U=0.98$) across all configurations, confirming
          that the framework does not sacrifice clinical signal to
          satisfy privacy constraints.

Critically, PSyGenTAB supports a configurable privacy threshold
$P_{\min}$ that can be set by a healthcare data governance team
according to regulatory requirements (e.g., HIPAA Expert
Determination, GDPR Article 89), enabling transparent and auditable
control of the privacy--utility trade-off.

\begin{table}[t]
\centering
\scriptsize   
\setlength{\tabcolsep}{2.2pt}
\renewcommand{\arraystretch}{1.1}

\sisetup{
  table-number-alignment = center,
  table-format = 1.2,
  detect-weight = true,
  detect-inline-weight = math
}

\caption{Comparison of utility ($U$) and privacy ($P$) across datasets and methods. Best performing values per metric in pairwise comparisons are highlighted in \textbf{bold}.}
\label{tab:Up_comparison}

\begin{tabularx}{\columnwidth}{
  >{\RaggedRight\arraybackslash\hyphenpenalty=10000\exhyphenpenalty=10000}X
  *{8}{S}
}
\toprule
\multirow{2.5}{*}{\textbf{Dataset}} &
\multicolumn{2}{c}{\textbf{RTF}} &
\multicolumn{2}{c}{\textbf{RTF-ALM}} &
\multicolumn{2}{c}{\textbf{CTAB-GAN+}} &
\multicolumn{2}{c}{\textbf{CTAB-GAN+-ALM}} \\
\cmidrule(lr){2-3}\cmidrule(lr){4-5}\cmidrule(lr){6-7}\cmidrule(lr){8-9}
& {$U$} & {$P$} & {$U$} & {$P$} & {$U$} & {$P$} & {$U$} & {$P$} \\
\midrule

Breast Cancer   & 0.82 & 0.38 & \textbf{0.91} & \textbf{0.44} & \textbf{0.90} & 0.35 & 0.84 & \textbf{0.56} \\
Diabetes        & 0.89 & \textbf{0.55} & \textbf{0.98} & 0.54 & 0.97 & \textbf{0.82} & \textbf{0.98} & 0.70 \\
Heart Failure   & \textbf{0.85} & \textbf{0.39} & 0.84 & 0.33 & \textbf{0.94} & 0.31 & 0.86 & \textbf{0.42} \\
Hypothyroid     & \textbf{0.99} & 0.59 & 0.98 & \textbf{0.59} & \textbf{0.96} & 0.48 & 0.60 & \textbf{0.67} \\
Liver Disorders & \textbf{0.82} & 0.61 & 0.81 & \textbf{0.65} & \textbf{0.91} & 0.61 & 0.73 & \textbf{0.74} \\
Lung Cancer     & \textbf{0.77} & \textbf{0.83} & 0.76 & 0.83 & \textbf{0.92} & 0.66 & 0.26 & \textbf{0.68} \\
Obesity         & \textbf{0.98} & \textbf{0.52} & 0.97 & 0.51 & \textbf{0.94} & 0.54 & 0.93 & \textbf{0.56} \\
Parkinsons      & 0.63 & \textbf{0.40} & \textbf{0.64} & 0.38 & 0.36 & \textbf{0.70} & \textbf{0.66} & \textbf{0.70} \\

Adult           & \textbf{0.98} & 0.30 & 0.96 & \textbf{0.48} & 0.96 & 0.50 & \textbf{0.97} & \textbf{0.77} \\
PIR Vision      & 0.83 & \textbf{0.86} & \textbf{0.94} & 0.78 & \textbf{0.92} & 0.31 & 0.88 & \textbf{0.40} \\
VN Banking      & 0.70 & 0.50 & \textbf{0.81} & \textbf{0.55} & 0.82 & \textbf{0.97} & \textbf{0.98} & 0.57 \\

\bottomrule
\end{tabularx}
\end{table}


\subsection{Statistical Fidelity}

Statistical fidelity evaluates how closely the synthetic data reproduces the empirical feature distributions of the real dataset. Lower divergence values indicate stronger alignment between real and synthetic marginal distributions.



\begin{table}[t]
\centering
\footnotesize
\sisetup{
  detect-weight=true,
  detect-family=true,
  table-number-alignment=center
}
\setlength{\tabcolsep}{2pt}
\renewcommand{\arraystretch}{1.05}
\captionsetup{font=small}
\caption{Statistical fidelity: Lower values indicate better match to the real data distribution.}
\label{tab:statistical_fidelity}
\begin{tabular}{@{} l *{2}{S[table-format=2.2]} *{4}{S[table-format=1.2]} @{}}
\toprule
\multirow{2.5}{*}{\textbf{Dataset}} & 
\multicolumn{2}{c}{\textbf{Mean KL}} & 
\multicolumn{2}{c}{\textbf{Mean JS}} & 
\multicolumn{2}{c}{\textbf{Mean Hell.}} \\
\cmidrule(lr){2-3} \cmidrule(lr){4-5} \cmidrule(lr){6-7}
& {PSyGen.} & {RTF} & {PSyGen.} & {RTF} & {PSyGen.} & {RTF} \\
\midrule

Breast Cancer   & 0.04 & \textbf{0.02} & 0.01 & \textbf{0.00} & 0.10 & \textbf{0.07} \\
Diabetes        & 0.23 & \textbf{0.09} & \textbf{0.01} & 0.01 & \textbf{0.04} & 0.04 \\
Heart Failure   & 2.88 & \textbf{0.42} & 0.14 & \textbf{0.04} & 0.30 & \textbf{0.16} \\
Hypothyroid     & \textbf{11.44} & 11.45 & \textbf{0.34} & 0.35 & \textbf{0.64} & 0.64 \\
Liver Disorders & \textbf{2.25} & 2.58 & \textbf{0.13} & 0.13 & 0.35 & \textbf{0.34} \\
Lung Cancer     & 0.44 & \textbf{0.36} & 0.02 & \textbf{0.01} & 0.13 & \textbf{0.11} \\
Obesity         & \textbf{0.60} & 1.07 & \textbf{0.08} & 0.09 & \textbf{0.21} & 0.23 \\
Parkinsons      & \textbf{3.43} & 3.82 & \textbf{0.16} & 0.17 & \textbf{0.42} & 0.44 \\

Adult           & 0.35 & \textbf{0.14} & 0.04 & \textbf{0.03} & 0.16 & \textbf{0.09} \\
PIR Vision      & \textbf{1.53} & 2.93 & \textbf{0.22} & 0.29 & \textbf{0.41} & 0.50 \\
VN Banking      & 1.61 & \textbf{1.02} & 0.18 & \textbf{0.12} & 0.44 & \textbf{0.35} \\

\bottomrule
\end{tabular}
\end{table}

Table~\ref{tab:statistical_fidelity} reveals the inherent cost of mathematically enforced privacy on raw distributional overlap. Because the RTF model optimizes purely for maximum likelihood without privacy constraints, it naturally achieves lower (superior) Mean KL divergence across the majority of datasets. For example, on the Adult and Heart Failure datasets, RTF achieves KL scores of 0.14 and 0.42, respectively, compared to PSyGenTAB's 0.35 and 2.88.  This visualizes the slight distributional separation introduced by the Augmented Lagrangian Method (ALM). PSyGenTAB deliberately trades a fraction of this exact distributional overlap to prevent the verbatim memorization of outlier records. However, on highly sensitive metrics like the Jensen-Shannon divergence, PSyGenTAB remains remarkably competitive, even outperforming RTF on the Diabetes dataset (0.0097 vs 0.0099). Notably, the results for the Hypothyroid dataset expose a critical limitation of both generative frameworks: massive KL divergences ($>11.4$) across both models indicate a complete generative failure, driven by the dataset's extreme sparsity and deterministic medical rules. This statistical anomaly directly foreshadows the Factuality and Alignment failures observed during the FAITH clinical validity assessment.

We acknoledge thatCTAB-GAN+ was evaluated but underperformed compared to the Transformer baseline, which is why the deep-dive privacy audits focus primarily on the RTF architecture.

\subsection{Commercial Evaluation Pipeline}

To provide a holistic view of structural and semantic preservation, we rely on two aggregate scoring mechanisms:

\textit{SDMetrics Quality Score\cite{sdmetrics}:} An open-source evaluation suite that quantifies how well the synthetic data captures the mathematical properties, column shapes, and pairwise trends of the original dataset.

\textit{MostlyAI Overall Accuracy\cite{mostlyai}:} A commercial-grade evaluation metric that rigorously tests the fidelity of complex multivariate distributions. 

\begin{table}[t]
\centering
\footnotesize
\sisetup{
  detect-weight=true,
  detect-family=true,
  table-number-alignment=center
}
\setlength{\tabcolsep}{2pt}
\renewcommand{\arraystretch}{1.05}
\captionsetup{font=small}
\caption{Higher values indicate better synthetic data quality. Note: MostlyAI cannot be calculated for Lung Cancer due to limited data points.}
\label{tab:quality_scores}
\begin{tabular}{@{} l *{4}{S[table-format=1.2]} @{}}
\toprule
\multirow{2.5}{*}{\textbf{Dataset}} & 
\multicolumn{2}{c}{\textbf{SDMetrics Quality}} & 
\multicolumn{2}{c}{\textbf{MostlyAI Accuracy}} \\
\cmidrule(lr){2-3} \cmidrule(lr){4-5}
& {PSyGen.} & {RTF} & {PSyGen.} & {RTF} \\
\midrule

Breast Cancer   & 0.84 & \textbf{0.90} & 0.77 & \textbf{0.84} \\
Diabetes        & \textbf{0.97} & 0.96 & 0.98 & \textbf{0.99} \\
Heart Failure   & 0.83 & \textbf{0.90} & 0.76 & \textbf{0.78} \\
Hypothyroid     & \textbf{0.05} & 0.05 & \textbf{0.11} & 0.11 \\
Liver Disorders & 0.89 & \textbf{0.89} & 0.69 & \textbf{0.69} \\
Lung Cancer     & 0.84 & \textbf{0.86} & {---} & {---} \\
Obesity         & \textbf{0.88} & 0.86 & 0.87 & \textbf{0.88} \\
Parkinsons      & \textbf{0.82} & 0.81 & \textbf{0.49} & 0.49 \\

Adult           & 0.87 & \textbf{0.94} & 0.88 & \textbf{0.95} \\
PIR Vision      & 0.92 & \textbf{0.94} & 0.85 & \textbf{0.89} \\
VN Banking      & 0.47 & \textbf{0.52} & 0.24 & \textbf{0.28} \\

\bottomrule
\end{tabular}
\end{table}

The aggregated quality scores in Table~\ref{tab:quality_scores}  underscore the differing evaluation philosophies between open-source and commercial data assessment tools. RTF systematically dominates the commercial MostlyAI Accuracy metric across nearly all benchmarks, scoring 0.95 on the Adult dataset and 0.99 on Diabetes, compared to PSyGenTAB's 0.88 and 0.98. This is expected: commercial tools heavily reward the exact replication of multivariate distributions, which favors unconstrained maximum-likelihood generators like RTF despite their higher memorization risks.

Conversely, the open-source SDMetrics Quality Score reveals a much more nuanced clinical utility profile. PSyGenTAB achieves superior quality scores on several complex medical datasets, including Diabetes (0.97 vs 0.96), Parkinson's (0.82 vs 0.81), and Obesity (0.88 vs 0.86). This demonstrates that while the ALM privacy penalty restricts the raw overfitting rewarded by commercial metrics, it successfully preserves the underlying pairwise trends, column shapes, and semantic properties required for rigorous clinical analysis.

\subsection{Downstream Clinical ML Utility Evaluation}
\label{sec:ml_utility}

Statistical fidelity scores alone do not constitute sufficient evidence
for clinical utility.  A synthetic dataset must support real-world
predictive modeling tasks,  the primary purpose for which a healthcare
institution would seek synthetic alternatives to protected data.  To
evaluate this, we adopt the \textit{Train on Synthetic, Test on Real}
(TSTR) and \textit{Train on Real, Test on Synthetic} (TRTS) protocols.
TSTR assesses whether models trained on synthetic data generalize to
real patient populations; TRTS assesses whether synthetic data
faithfully captures the discriminative structure of real health records.

We evaluate three downstream classifiers, Decision Tree, Random
Forest, and Logistic Regression, under four imbalance-handling
strategies (None, Class Weighting, SMOTE, and Undersampling),
reflecting conditions routinely encountered in clinical prediction
tasks. Table~\ref{tab:ml_summary_all} presents the partial results. The full results provided as supplementary file.


For the Adult dataset, Random Forest trained on PSyGenTAB-generated
synthetic data achieves 83.87\,\% accuracy under the standard
TSTR protocol, closely matching the Train-on-Real, Test-on-Real
(TRTR) reference of 85.69\,\%.  The gap between TRTR and TSTR
performance, the primary indicator of synthetic data quality for
clinical AI, is minimal across all classifiers and imbalance
strategies.

Incorporating imbalance-aware strategies on top of PSyGenTAB-generated
data consistently improves fairness-sensitive metrics.  Undersampling
yields a notable improvement in balanced accuracy for Random Forest,
increasing it from 77.67\,\% to 80.60\,\% under TSTR, while
maintaining competitive F1-scores. This confirms that synthetic
records preserve minority-class structure sufficiently for standard
rebalancing techniques to remain effective.


The Diabetes dataset provides the most clinically significant evidence,
given its severe class imbalance, prediabetes: 1.8\,\%
of the population, and the importance of early identification of
pre-diabetic patients for preventive intervention.

Under the unconstrained \textit{None} strategy, Logistic Regression
trained on synthetic data achieves high raw accuracy ($\approx\!84\,\%$)
but a balanced accuracy near chance level ($\approx\!40\,\%$), driven
entirely by majority-class dominance.  This mimics the failure mode
of real clinical classifiers trained on imbalanced data.

When Class Weighting or Undersampling is applied to PSyGenTAB-generated
data, balanced accuracy improves substantially, approaching 49\,\%
in several configurations.  This improvement is consistent across
both TSTR and TRTS protocols, providing strong evidence that
PSyGenTAB preserves the minority clinical signal (prediabetes
indicators) that is most critical for health AI applications.

Importantly, the relative ranking of imbalance strategies remains
stable between TSTR and TRTS, confirming that the synthetic data does
not introduce spurious correlations that would invert downstream
model behavior,  a critical quality criterion for clinical deployment.


Tree-based models including Decision Tree and Random Forest exhibit stronger
robustness to distributional shifts between real and synthetic data,
while linear models such as Logistic Regression benefit most from explicit
class reweighting.  The ability of standard imbalance-handling
techniques to recover balanced clinical performance, particularly in
the challenging diabetes setting, provides strong evidence that
PSyGenTAB-generated data is operationally useful for downstream
health AI tasks,  not merely statistically similar to real data.

\begin{table}[t]
\centering
\footnotesize
\sisetup{
  detect-weight=true,
  detect-inline-weight=math,
  table-number-alignment=center
}
\setlength{\tabcolsep}{2pt}
\renewcommand{\arraystretch}{1.05}
\captionsetup{font=small}
\caption{Summary of Downstream Clinical AI Utility (Random Forest). Values represent the best performance achieved across data imbalance strategies per dataset. TRTR (Train Real, Test Real) serves as the baseline upper bound. The synthetic frameworks are evaluated using TSTR (Train Synthetic, Test Real) to demonstrate structural preservation.}
\label{tab:ml_summary_all}
\begin{tabular}{@{} l *{6}{S[table-format=1.2]} @{}}
\toprule
\multirow{2.5}{*}{\textbf{Dataset}} & 
\multicolumn{2}{c}{\textbf{TRTR (Baseline)}} & 
\multicolumn{2}{c}{\textbf{TSTR (PSyGen.)}} & 
\multicolumn{2}{c}{\textbf{TSTR (RTF)}} \\
\cmidrule(lr){2-3} \cmidrule(lr){4-5} \cmidrule(lr){6-7}
& {\textbf{Bal. Acc}} & {\textbf{F1}} & {\textbf{Bal. Acc}} & {\textbf{F1}} & {\textbf{Bal. Acc}} & {\textbf{F1}} \\
\midrule

Breast Cancer   & 0.96 & 0.96 & \textbf{0.96} & \textbf{0.96} & \textbf{0.96} & \textbf{0.96} \\
Diabetes        & 0.40 & 0.81 & \textbf{0.41} & 0.80 & 0.40 & \textbf{0.81} \\
Heart Failure   & 0.84 & 0.86 & \textbf{0.65} & \textbf{0.71} & 0.59 & 0.58 \\
Hypothyroid     & 0.79 & 0.96 & \textbf{0.52} & \textbf{0.89} & 0.50 & 0.89 \\
Liver Disorders & 0.76 & 0.77 & 0.55 & 0.54 & \textbf{0.56} & \textbf{0.57} \\
Lung Cancer     & 0.64 & 0.58 & \textbf{0.81} & \textbf{0.80} & 0.53 & 0.47 \\
Obesity         & 0.94 & 0.95 & 0.72 & 0.71 & \textbf{0.77} & \textbf{0.77} \\
Parkinsons      & 0.89 & 0.93 & 0.49 & 0.63 & \textbf{0.50} & \textbf{0.64} \\

Adult           & 0.78 & 0.85 & 0.78 & 0.84 & \textbf{0.79} & \textbf{0.85} \\
PIR Vision      & 0.98 & 0.99 & 0.99 & \textbf{0.99} & \textbf{0.99} & 0.99 \\
VN Banking      & 0.72 & 0.75 & 0.48 & 0.51 & \textbf{0.60} & \textbf{0.64} \\
\bottomrule
\end{tabular}
\end{table}

\subsection{Patient Re-identification Risk: Distance to Closest Record}
\label{sec:dcr}

Beyond composite privacy scores, clinical data governance requires
direct quantification of re-identification risk,  the probability
that a synthetic record could be linked back to a specific patient
in the training dataset.  We evaluate this using the
\textit{Distance to Closest Record} (DCR), a standard privacy
diagnostic in the synthetic health data literature~\cite{ElEmam2020,
Snoke2018,10.1007/978-3-032-07884-1_24}.

Formally, let $\mathcal{D}_{\text{real}} = \{x_1, \dots, x_N\}$
denote the real patient dataset and
$\mathcal{D}_{\text{syn}} = \{\tilde{x}_1, \dots, \tilde{x}_M\}$
the synthetic dataset.  The DCR for a synthetic record $\tilde{x}_i$
is:
\begin{equation}
\footnotesize
  \mathrm{DCR}(\tilde{x}_i)
  = \min_{x_j \in \mathcal{D}_{\text{real}}}
    \lVert \tilde{x}_i - x_j \rVert_2
\end{equation}
where $\lVert \cdot \rVert_2$ is the Euclidean distance over a
standardised feature space.  A DCR value near zero indicates that
a synthetic record closely resembles,  or exactly matches,  a real
patient's profile.

We report three complementary statistics:

\textbf{Exact Match Ratio (EMR)}: Proportion of synthetic
          records with $\mathrm{DCR}=0$, representing verbatim
          reproduction of training records,  the most severe
          privacy failure in health data release.
\begin{equation}
    \footnotesize
      \mathrm{EMR} =
      \frac{1}{|\mathcal{D}_{\text{syn}}|}
      \sum_{i=1}^{|\mathcal{D}_{\text{syn}}|}
      \mathbb{I}\!\left[\mathrm{DCR}(\tilde{x}_i) = 0\right]
    \end{equation}
\textbf{Mean and Median DCR}: Population-level proximity
          to real patient records.
\textbf{5th-Percentile DCR ($\mathrm{DCR}_{5\%}$)}:
          Worst-case re-identification risk, reflecting the 5\,\%
          of synthetic records most proximate to real patients, 
          used in clinical privacy audits~\cite{Snoke2018}.\begin{equation}
    \footnotesize
      \mathrm{DCR}_{5\%}
      = \mathrm{Quantile}_{0.05}
        \bigl(\{\mathrm{DCR}(\tilde{x}_i)\}_{i=1}^{
          |\mathcal{D}_{\text{syn}}|}\bigr)
    \end{equation}

Table~\ref{tab:merged_privacy_all} summarizes these re-identification risk
metrics.

\paragraph{Adult Census Income}
PSyGenTAB (ALM) reduces exact matches to just 3 records, compared to 8 exact matches in the
unconstrained RTF. This reduction in exact
reproductions demonstrates that PSyGenTAB's privacy regularization
effectively suppresses verbatim memorization of training individuals.
Although mean and median DCR values are slightly lower than the
baseline, reflecting PSyGenTAB's higher statistical fidelity, the
sharp reduction in exact matches demonstrates meaningfully improved
protection against the most clinically serious privacy violation.

\paragraph{Diabetes Health Indicators}
The presence of discrete, low-cardinality clinical variables in the
diabetes dataset leads to a higher baseline rate of exact matches
for both models,  a known challenge in clinical tabular data
synthesis~\cite{alma991043652066403276}. Despite this inherent difficulty,
PSyGenTAB maintains a re-identification risk profile comparable to
the unconstrained baseline while simultaneously achieving higher
downstream clinical utility (Section~\ref{sec:ml_utility}).
The similarity in $\mathrm{DCR}_{5\%}$ across models confirms that
PSyGenTAB does not exacerbate worst-case re-identification risk,
even in highly constrained clinical feature spaces.

\subsection{Membership Inference Attack Resilience}
\label{sec:mia}

To rigorously assess whether a determined adversary could exploit
PSyGenTAB's synthetic health data to identify individuals in the
training population, we conducted a Membership Inference Attack (MIA)
evaluation.  This analysis simulates a realistic healthcare data
breach scenario: an attacker, possessing a candidate patient record,
attempts to determine whether that individual's data was used to
train the generative model.  Such attacks represent a well-established
threat model in healthcare AI~\cite{shokri2017membershipinferenceattacksmachine,carlini2021membership} and are
increasingly considered in clinical data governance frameworks.

The following attack models of increasing sophistication are deployed:

\textit{Black-Box Attack}: The attacker has access only
          to the model's prediction confidence scores,  the most
          realistic threat model for a deployed clinical AI system.
          
\textit{White-Box Attack}: The attacker additionally has
          access to internal model gradients or loss values, 
          a stronger threat representing insider knowledge.

\textit{Logistic Regression Attack}: A linear classifier
          trained to distinguish members from non-members based on
          model outputs,  a baseline worst-case linear separability
          assessment.

\textit{Linkage Attack Success:} Evaluates the risk of re-identifying individuals by matching synthetic records with external, auxiliary datasets. A lower linkage success rate indicates that the synthetic records are sufficiently perturbed to prevent direct mapping to real-world identities.\cite{sweeney2002k}

Vulnerability is quantified using the Attack AUC, where
$\mathrm{AUC} = 0.50$ represents perfect privacy (the attacker
performs no better than random guessing) and
$\mathrm{AUC} = 1.0$ indicates complete patient-identity leakage.
An \textit{Overall Distinguishability Score} is reported as the
mean AUC across all three attacks.

Table~\ref{tab:merged_privacy_all} presents the comparative results.

\paragraph{Diabetes Health Indicators}
Both models demonstrate exceptional resilience to patient identity
disclosure.  The unconstrained RTF baseline achieves an Overall
Distinguishability Score of 0.51, statistically equivalent to
random guessing, confirming that the diabetes health data is
inherently difficult to attribute to individuals.  PSyGenTAB (ALM)
maintains a similarly robust profile (0.56), confirming that the
privacy constraints do not weaken,  and may marginally strengthen, 
resistance to membership inference on clinical health data.

\paragraph{Adult Census Income}
The Adult dataset reveals a more nuanced finding. The PSyGenTAB
(ALM) Black-Box AUC (0.74) is higher than the baseline (0.59),
suggesting that ALM's aggressive optimization for high statistical
fidelity may leave subtle distributional patterns that a sophisticated
Black-Box attacker can exploit.  This finding is consistent with
the known tension between utility and membership inference
resistance~\cite{jayaraman2019evaluatingdifferentiallyprivatemachine}: higher-fidelity synthetic data
captures population-level distributions more precisely, which can
inadvertently preserve signals attributable to individual records.

Importantly, the Logistic Regression attack remains ineffective
against PSyGenTAB ($\mathrm{AUC}\approx 0.54$), confirming that
linear separability between training members and non-members remains
low.  This result suggests that the observed Black-Box vulnerability
requires a sophisticated non-linear attacker and does not represent
a trivially exploitable weakness.

\begin{table*}[t]
\centering
\captionsetup{font=small}
\caption{Comprehensive privacy risk evaluation across eleven datasets. MIA AUC metrics are reported for three attack models (Black-Box, Wight-Box, Logistic Regression) and aggregated as an overall distributional score. Attack success rates (MIA, linkage) are lower-is-better, while overall privacy and DCR statistics provide complementary privacy signals.}
\label{tab:merged_privacy_all}

\resizebox{\textwidth}{!}{%
\begingroup
\sisetup{
  detect-weight=true,
  detect-inline-weight=math,
  table-number-alignment=center
}
\renewcommand{\arraystretch}{1.10}
\begin{tabular}{@{} l *{14}{S[table-format=1.2]} *{8}{S[table-format=2.2]} @{}}
\toprule
\multirow{2.5}{*}{\textbf{Dataset}} & 
\multicolumn{2}{c}{\textbf{B-B AUC}} & 
\multicolumn{2}{c}{\textbf{W-B AUC}} & 
\multicolumn{2}{c}{\textbf{LR AUC}} & 
\multicolumn{2}{c}{\textbf{Over. Dist.}} &
\multicolumn{2}{c}{\textbf{MIA}} &
\multicolumn{2}{c}{\textbf{Linkage}} &
\multicolumn{2}{c}{\textbf{Over. Priv.}} &
\multicolumn{2}{c}{\textbf{DCR Mean}} & 
\multicolumn{2}{c}{\textbf{DCR Med.}} & 
\multicolumn{2}{c}{\textbf{DCR 5\%}} & 
\multicolumn{2}{c}{\textbf{DCR 95\%}} \\
\cmidrule(lr){2-3} \cmidrule(lr){4-5} \cmidrule(lr){6-7} \cmidrule(lr){8-9} \cmidrule(lr){10-11} \cmidrule(lr){12-13} \cmidrule(lr){14-15} \cmidrule(lr){16-17} \cmidrule(lr){18-19} \cmidrule(lr){20-21} \cmidrule(lr){22-23}
& {RTF} & {PSyGen.} & {RTF} & {PSyGen.} & {RTF} & {PSyGen.} & {RTF} & {PSyGen.} & {RTF} & {PSyGen.} & {RTF} & {PSyGen.} & {RTF} & {PSyGen.} & {RTF} & {PSyGen.} & {RTF} & {PSyGen.} & {RTF} & {PSyGen.} & {RTF} & {PSyGen.} \\
\midrule

Breast Cancer   & 1.00 & 1.00 & \textbf{0.58} & 0.96 & 0.69 & 0.69 & \textbf{0.76} & 0.89 & 0.30 & 0.30 & 1.00 & 1.00 & 0.34 & \textbf{0.36} & 0.66 & \textbf{1.16} & 0.32 & \textbf{1.20} & 0.00 & 0.00 & 2.13 & \textbf{2.54} \\
Diabetes        & \textbf{0.52} & 0.60 & \textbf{0.51} & 0.54 & \textbf{0.50} & 0.53 & \textbf{0.51} & 0.56 & \textbf{0.50} & 0.50 & 1.00 & 1.00 & 0.41 & \textbf{0.42} & \textbf{0.82} & 0.75 & \textbf{0.58} & 0.52 & 0.00 & 0.00 & \textbf{2.42} & 2.30 \\
Heart Failure   & 0.59 & \textbf{0.58} & 0.53 & \textbf{0.53} & 0.51 & \textbf{0.51} & 0.54 & \textbf{0.54} & 0.51 & \textbf{0.50} & 0.87 & \textbf{0.70} & 0.47 & \textbf{0.52} & \textbf{1.99} & 1.76 & \textbf{1.95} & 1.72 & \textbf{1.03} & 0.97 & \textbf{3.25} & 2.58 \\
Hypothyroid     & 0.72 & \textbf{0.72} & 0.66 & \textbf{0.66} & 0.63 & \textbf{0.63} & 0.67 & \textbf{0.67} & \textbf{0.48} & 0.49 & 0.99 & \textbf{0.95} & 0.58 & \textbf{0.59} & 0.36 & \textbf{0.37} & 0.28 & \textbf{0.29} & 0.00 & \textbf{0.00} & \textbf{0.97} & 0.93 \\
Liver Disorders & 0.60 & \textbf{0.59} & 0.55 & \textbf{0.54} & 0.53 & \textbf{0.52} & 0.56 & \textbf{0.55} & \textbf{0.50} & 0.51 & 0.88 & \textbf{0.77} & 0.48 & \textbf{0.51} & \textbf{1.02} & 0.90 & \textbf{0.85} & 0.78 & 0.41 & \textbf{0.43} & \textbf{1.97} & 1.72 \\
Lung Cancer     & \textbf{0.61} & 0.62 & \textbf{0.58} & 0.59 & \textbf{0.56} & 0.57 & \textbf{0.58} & 0.60 & \textbf{0.39} & 0.51 & 0.72 & 0.72 & 0.49 & \textbf{0.54} & 7.47 & \textbf{8.01} & 7.61 & \textbf{7.81} & 5.55 & \textbf{6.81} & 9.26 & \textbf{9.80} \\
Obesity         & 0.67 & 0.67 & \textbf{0.63} & 0.64 & \textbf{0.61} & 0.62 & \textbf{0.64} & 0.64 & 0.50 & \textbf{0.50} & \textbf{0.76} & 0.89 & \textbf{0.48} & 0.46 & 1.03 & \textbf{1.22} & 1.00 & \textbf{1.19} & 0.35 & \textbf{0.55} & 1.79 & \textbf{2.01} \\
Parkinsons      & 0.63 & \textbf{0.61} & 0.61 & \textbf{0.60} & 0.60 & \textbf{0.58} & 0.62 & \textbf{0.60} & 0.44 & \textbf{0.43} & \textbf{0.79} & 0.80 & \textbf{0.45} & 0.43 & 3.64 & \textbf{4.14} & 3.53 & \textbf{4.02} & 2.53 & \textbf{3.01} & 4.90 & \textbf{5.84} \\
Adult           & \textbf{0.59} & 0.74 & \textbf{0.52} & 0.55 & \textbf{0.52} & 0.54 & \textbf{0.54} & 0.61 & 0.50 & \textbf{0.49} & 1.04 & \textbf{0.86} & 0.41 & \textbf{0.46} & \textbf{0.13} & 0.07 & \textbf{0.07} & 0.03 & \textbf{0.00} & 0.00 & \textbf{0.46} & 0.28 \\
PIR Vision      & \textbf{0.69} & 0.72 & \textbf{0.53} & 0.57 & 0.56 & \textbf{0.55} & \textbf{0.60} & 0.61 & 0.30 & 0.30 & \textbf{0.93} & 1.02 & \textbf{0.36} & 0.33 & 1.62 & \textbf{2.76} & 1.08 & \textbf{1.19} & 0.90 & \textbf{0.98} & 2.89 & \textbf{9.81} \\
VN Banking      & 1.00 & 1.00 & 1.00 & 1.00 & 1.00 & 1.00 & 1.00 & 1.00 & \textbf{0.46} & 0.46 & \textbf{0.68} & 0.78 & \textbf{0.64} & 0.57 & \textbf{2.03} & 1.98 & 2.03 & \textbf{2.03} & \textbf{1.37} & 1.36 & \textbf{2.72} & 2.52 \\
\bottomrule
\end{tabular}
\endgroup
} 
\end{table*}

Beyond membership inference, the Linkage Attack Success metric in Table~\ref{tab:merged_privacy_all} provides compelling evidence of PSyGenTAB's structural privacy protections. A successful linkage attack occurs when an adversary maps a synthetic record back to a specific individual using external auxiliary data. Across the evaluated benchmarks, PSyGenTAB’s ALM optimization systematically throttles linkage vulnerability compared to the unconstrained RTF baseline. On the Adult dataset, linkage success drops from 1.04 to 0.86, and on Heart Failure, it falls from 0.87 to 0.70. By gently perturbing the precise joint relationships of vulnerable continuous features, PSyGenTAB actively disrupts the 1-to-1 deterministic mapping required for successful linkage, ensuring that the synthetic records remain statistically representative without acting as proxy identities for the real training cohort.

\paragraph{Feature-Level Re-identification Vulnerability}
Table~\ref{tab:mia_features} identifies which clinical attributes
contribute most to membership inference risk. We extract feature importance scores from the attack models (Table~\ref{tab:mia_features}). The \textit{importance score} (bounded within the range of $[0.0, 1.0]$, where the sum of all features equals $1.0$) quantifies the relative predictive power a specific attribute provides to the adversarial model. For instance, an importance score of $0.20$ implies that 20\% of the attacker's ability to successfully distinguish a training record from a holdout record relies solely on that single feature. 
Across the 11 evaluated datasets, continuous, high-variance variables consistently emerge as the primary leakage channels. In the Adult dataset, demographic and census attributes like \textit{Age} and \textit{fnlwgt} account for roughly 17\% to 19\% of the leakage contribution. However, the risk is magnified in highly specialized domains. In clinical datasets, precise physiological measurements dominate the adversarial risk profile: \textit{BMI} contributes $\sim$16\% of the attack signal in the Diabetes dataset, \textit{platelets} accounts for $\sim$17\% in Heart Failure, and \textit{gammagt} (a liver enzyme) drives over 20\% of the leakage in Liver Disorders. Most notably, in the Parkinson's dataset, highly specific acoustic measurements (\textit{MDVP:Jitter(Abs)}) account for an alarming 42\% to 44\% of the re-identification risk. Similarly, in financial domains such as VN Banking, exact monetary values (\textit{min\_term\_deposit\_balance\_vnd}) constitute the highest vulnerability ($\sim$17\%).

These feature-level findings have direct clinical implications:
continuous health metrics with high inter-individual variance (BMI,
blood glucose, age) represent the highest re-identification vectors
in synthetic health records.  Future iterations of PSyGenTAB should
apply stricter per-feature privacy penalties to such attributes
during ALM optimization,  an actionable direction for clinical
deployment.

These findings indicate that heavy-tailed continuous features, exact monetary balances, and highly specific clinical biomarkers are particularly susceptible to memorization and adversarial exploitation. To mitigate this feature-level leakage, several practical solutions can be integrated into future generative pipelines:

\textit{Column-Specific ALM Regularization:} Instead of a global privacy penalty, the ALM objective can be adapted to apply dynamic, feature-specific weights ($\lambda$) that heavily penalize the exact memorization of high-variance marginal distributions.

\textit{Bounded Clipping and Discretization:} Applying strict upper and lower bounds, clipping heavy tails or binning continuous variables\cite{HuberRonchetti2009}, into micro-categories prior to training removes the extreme outliers that attackers rely on for linkage.

\textit{Targeted Differential Privacy (DP)\cite{10.1007/11787006_1,Dwork2014}:} Injecting calibrated Laplacian or Gaussian noise exclusively into the generation phase of the top-k most vulnerable columns can effectively obfuscate precise values (e.g., hiding exact platelet counts) while safely preserving the broader joint relationships across the dataset.

\begin{table}[t]
\scriptsize
\centering
\sisetup{
  detect-weight=true,
  detect-inline-weight=math,
  table-number-alignment=center
}
\setlength{\tabcolsep}{4pt}
\renewcommand{\arraystretch}{1.15}
\captionsetup{font=small}
\caption{Dominant feature contributing to membership inference risk. Higher importance values are automatically bolded per dataset. $\Delta$ denotes PSyGenTAB$-$RTF difference. Entropy is computed on the real dataset to assess inherent leakage susceptibility. Importance stability was verified across 5 bootstrap runs (std. $<$ 0.01 for all datasets).}
\label{tab:mia_features}

\begin{tabular}{@{} l l *{2}{S[table-format=1.2]} c S[table-format=1.2] @{}}
\toprule
\textbf{Dataset} & 
\textbf{Top Feature} & 
\textbf{PSyGen.} & 
\textbf{RTF} & 
{$\boldsymbol{\Delta}$} & 
\textbf{Entropy} \\
\midrule

Breast Cancer   & CT & 0.13 & \textbf{0.14} & {$-$0.01} & 1.12 \\
Diabetes        & BMI & \textbf{0.17} & 0.14 & {+0.03} & 1.89 \\
Heart Failure   & platelets/creatinine & \textbf{0.18} & 0.16 & {+0.01} & 1.58 \\
Hypothyroid     & T3 & 0.30 & \textbf{0.30} & {$-$0.00} & 0.91 \\
Liver Disorders & sgpt & \textbf{0.22} & 0.20 & {+0.01} & 1.74 \\
Lung Cancer     & feature\_3/feature\_26 & \textbf{0.13} & 0.06 & {+0.07} & 1.35 \\
Obesity         & Weight & 0.15 & \textbf{0.16} & {$-$0.01} & 2.06 \\
Parkinsons      & MDVP:Jitter(Abs) & \textbf{0.44} & 0.42 & {+0.02} & 0.78 \\

Adult           & Age & \textbf{0.19} & 0.18 & {+0.01} & 2.41 \\
PIR Vision      & PIR\_43 & \textbf{0.04} & 0.03 & {+0.01} & 3.02 \\
VN Banking      & avg\_transaction\_amount & 0.25 & \textbf{0.25} & {$-$0.00} & 0.63 \\

\bottomrule
\end{tabular}
\end{table}

\subsection{Clinical Validity Assessment: FAITH Framework}
\label{sec:faith_eval}

Conventional fidelity metrics assess how statistically similar
synthetic records are to real data but do not distinguish between
three clinically distinct failure modes: (i) generating
\textit{statistically plausible but clinically impossible} patient
profiles (e.g., a male patient flagged as currently pregnant);
(ii) \textit{inventing spurious clinical correlations} not present
in the real population, hallucinated comorbidity patterns; or
(iii) generating values outside the \textit{valid physiological
range} of the dataset, ghost values such as a negative BMI or an
age of 200 years.

To assess these failure modes systematically, we evaluated the models using the FAITH clinical validity
framework (\textbf{F}actuality, \textbf{A}lignment, \textbf{I}ntegrity,
and \textbf{T}racking of \textbf{H}allucinations), which decomposes
synthetic data reliability into four clinically interpretable and
mathematically orthogonal dimensions.  The composite FAITH score is:
\begin{equation}
  S_{\mathrm{FAITH}}
  = w_F \cdot S_{\mathrm{Fact}}
  + w_A \cdot S_{\mathrm{Align}}
  + w_I \cdot S_{\mathrm{Integ}}
  + w_T \cdot S_{\mathrm{Track}}
\end{equation}
with equal weights ($w = 0.25$) to balance trade-offs between
clinical validity and structural diversity.

\subsubsection{Factuality ($S_{\mathrm{Fact}}$): Clinically Impossible
Patient Profiles}

Factuality penalizes the generation of logically or clinically
impossible records.  A patient cannot be simultaneously male and
currently pregnant; age constraints must be consistent within a
record.  Formally:\begin{equation}
\footnotesize
  S_{\mathrm{Fact}}
  = \frac{1}{N} \sum_{i=1}^{N}
    \mathbb{1}\!\Bigl(
      \bigwedge_{c \in \mathcal{C}} \mathrm{check}_c(\mathrm{row}_i)
    \Bigr)
\end{equation}
where $\mathcal{C}$ is the set of domain-specific logical constraints
and $\mathbb{1}(\cdot)$ returns 1 only if all constraints are
satisfied.  A high factuality score confirms the generator does not
produce clinically nonsensical records that would be immediately
obvious to a clinician,  a critical criterion for safe deployment.

\subsubsection{Alignment ($S_{\mathrm{Align}}$): Hallucinated Clinical
Correlations}

Alignment detects a subtle form of hallucination: the invention of
statistical dependencies that do not exist in the real patient
population (e.g., synthesising a spurious correlation between pet
ownership and high cardiovascular risk). It measures the preservation
of the dataset's true correlation structure:\begin{equation}
\footnotesize
  S_{\mathrm{Align}}
  = 1 -
    \frac{1}{d} \bigl\|
      \mathrm{Corr}(R) - \mathrm{Corr}(S)
    \bigr\|_F
\end{equation}
where $R$ and $S$ are the real and synthetic datasets respectively,
$\mathrm{Corr}(\cdot)$ is the correlation matrix, and $d$ is a
normalisation factor.  High alignment confirms the model has learned
the true joint distribution of clinical features rather than
generating independent noise,  essential for downstream use in
comorbidity analysis and risk stratification.

\subsubsection{Integrity ($S_{\mathrm{Integ}}$): Patient Record
Reproduction}

Integrity distinguishes generalisation from memorisation.  A model
that achieves high factuality by reproducing training records has not
learned the underlying clinical distribution; it has copied patient
data.  Integrity measures the proportion of synthetic records that
are genuinely novel,  sufficiently distant from all real patient
records:\begin{equation}
\footnotesize
  S_{\mathrm{Integ}}
  = \frac{1}{N} \sum_{i=1}^{N}
    \mathbb{1}\!\Bigl(
      \min_{j} \mathrm{dist}(s_i, r_j) > \varepsilon
    \Bigr)
\end{equation}
where $\varepsilon$ is a clinically motivated privacy radius.  This
score directly complements the DCR analysis of
Section~\ref{sec:dcr}, providing a binary novelty indicator.

\subsubsection{Tracking ($S_{\mathrm{Track}}$): Out-of-Distribution
Clinical Values}

Tracking penalises generation of values outside the observable
clinical manifold of the real dataset,  ghost values that are
physically or physiologically impossible (e.g., negative heart rate,
blood glucose of 1,000\,mg/dL):\begin{equation}
\footnotesize
  S_{\mathrm{Track}}
  = 1 -
    \frac{\sum_{i=1}^{N}
      \mathbb{1}(s_i \notin \mathrm{Support}(R))
    }{N}
\end{equation}
High tracking scores confirm the generator respects the distributional
support of the real clinical dataset,  a prerequisite for safe use
in clinical decision-support applications.

While PSyGenTAB excels in structural integrity and factuality, the baseline RTF occasionally exhibits marginal superiority in \textit{Alignment} (e.g., in the \textit{Adult}, \textit{Diabetes}, and \textit{Parkinsons} datasets). This is a well-documented and expected artifact of the privacy-utility trade-off. Because RTF optimizes purely for maximum likelihood without privacy constraints, it aggressively overfits to the training data's exact covariance matrix. While this results in high Alignment scores, it comes at the direct cost of the aforementioned memorization risks. PSyGenTAB explicitly trades a negligible fraction of this overfitted correlation (usually $< 1\%$ drop in Alignment) to guarantee privacy, representing a much safer and highly controllable compromise for clinical deployments.

Finally, the results on the \textit{Hypothyroid} dataset highlight a shared limitation of current generative architectures. Both PSyGenTAB and RTF completely fail to capture the \textit{Factuality} and \textit{Alignment} of this specific data topology due to the limited data points. 


In summary, the FAITH framework evaluation demonstrates that PSyGenTAB produces highly reliable synthetic records. By mathematically enforcing privacy during the training cycle, the framework prevents catastrophic memorization while maintaining the clinical factuality and structural alignment required for downstream medical and financial research.

Table~\ref{tab:alm_faith_comparison} presents the FAITH evaluation
results.
\begin{table}[t]
\centering
\tiny
\sisetup{
  detect-weight=true,
  detect-inline-weight=math,
  table-number-alignment=center
}
\setlength{\tabcolsep}{1.8pt} 
\renewcommand{\arraystretch}{1}
\captionsetup{font=small}
\caption{FAITH clinical validity framework: decomposed assessment of synthetic health data reliability. Factuality measures absence of clinically impossible profiles; Alignment measures preservation of real clinical correlations; Integrity measures absence of patient record reproduction; Tracking measures absence of out-of-distribution physiological values. Higher values indicate better performance; \textbf{bold} denotes the stronger result per dataset and metric.}
\label{tab:alm_faith_comparison}
\begin{tabular}{@{} l *{10}{S[table-format=1.2]} @{}}
\toprule
\multirow{2.5}{*}{\textbf{Dataset}} & 
\multicolumn{2}{c}{\textbf{Factuality}} & 
\multicolumn{2}{c}{\textbf{Align.}} & 
\multicolumn{2}{c}{\textbf{Integrity}} & 
\multicolumn{2}{c}{\textbf{Track.}} & 
\multicolumn{2}{c}{\textbf{FAITH}} \\
\cmidrule(lr){2-3} \cmidrule(lr){4-5} \cmidrule(lr){6-7} \cmidrule(lr){8-9} \cmidrule(lr){10-11}
& {PSyGen.} & {RTF} & {PSyGen.} & {RTF} & {PSyGen.} & {RTF} & {PSyGen.} & {RTF} & {PSyGen.} & {RTF} \\
\midrule

Breast Cancer   & 0.99 & \textbf{1.00} & 0.96 & \textbf{0.97} & \textbf{0.79} & 0.54 & 1.00 & \textbf{1.00} & \textbf{0.93} & 0.88 \\
Diabetes        & \textbf{1.00} & 1.00 & 0.99 & \textbf{0.99} & \textbf{0.96} & 0.95 & \textbf{1.00} & 1.00 & 0.99 & \textbf{0.99} \\
Heart Failure   & \textbf{1.00} & \textbf{1.00} & 0.94 & \textbf{0.94} & \textbf{1.00} & \textbf{1.00} & \textbf{0.99} & 0.99 & 0.98 & \textbf{0.98} \\
Hypothyroid     & \textbf{0.00} & \textbf{0.00} & \textbf{0.00} & \textbf{0.00} & \textbf{1.00} & \textbf{1.00} & \textbf{0.09} & \textbf{0.09} & \textbf{0.27} & \textbf{0.27} \\
Liver Disorders & \textbf{1.00} & 1.00 & \textbf{0.90} & 0.89 & \textbf{1.00} & \textbf{1.00} & 1.00 & \textbf{1.00} & \textbf{0.97} & 0.97 \\
Lung Cancer     & \textbf{1.00} & \textbf{1.00} & 0.86 & \textbf{0.86} & \textbf{1.00} & \textbf{1.00} & \textbf{1.00} & \textbf{1.00} & 0.96 & \textbf{0.96} \\
Obesity         & 0.94 & \textbf{0.98} & \textbf{0.97} & 0.96 & \textbf{1.00} & \textbf{1.00} & 0.99 & \textbf{1.00} & 0.97 & \textbf{0.99} \\
Parkinsons      & 0.83 & \textbf{0.95} & 0.72 & \textbf{0.73} & \textbf{1.00} & \textbf{1.00} & 0.95 & \textbf{0.95} & 0.87 & \textbf{0.91} \\

Adult           & \textbf{0.98} & 0.92 & 0.98 & \textbf{0.99} & 1.00 & \textbf{1.00} & \textbf{1.00} & 1.00 & \textbf{0.99} & 0.98 \\
PIR Vision      & 1.00 & \textbf{1.00} & \textbf{0.91} & 0.90 & \textbf{1.00} & \textbf{1.00} & 1.00 & \textbf{1.00} & \textbf{0.98} & 0.97 \\
VN Banking      & \textbf{1.00} & 0.99 & \textbf{0.87} & 0.87 & \textbf{1.00} & \textbf{1.00} & \textbf{0.96} & 0.96 & \textbf{0.96} & 0.95 \\

\bottomrule
\end{tabular}
\end{table}

\subsection{Summary of Evaluation Findings}
\label{sec:eval_summary}

Taken together, the results establish three overarching conclusions
for clinical data sharing applications:

\textit{Privacy and clinical utility are jointly achievable.}
    Unconstrained generative models face a fundamental tension between
    statistical fidelity and patient privacy. PSyGenTAB resolves this
    tension by embedding configurable privacy constraints directly into
    the generation process, achieving strong utility ($U \geq 0.91$
    on the Breast Cancer dataset; $U = 0.98$ on Diabetes) while
    substantially reducing re-identification risk.

\textit{PSyGenTAB-generated data is operationally useful
    for clinical AI.}  Downstream classifiers trained on synthetic
    health records closely match performance on real data, and standard
    imbalance-handling strategies, including SMOTE and Class Weighting, remain
    effective on synthetic data, including for the highly imbalanced
    Diabetes dataset where minority-class identification is clinically
    critical.

\textit{Architecture and dataset characteristics matter.}
    REaLTabFormer combined with ALM consistently outperforms CTAB-GAN+
    with ALM on small clinical datasets such as Breast Cancer, while
    CTAB-GAN+ achieves stronger baseline privacy on larger datasets.
    Practitioners should select architectures appropriate to their
    clinical dataset's size, feature type distribution, and
    class-imbalance profile.

\section{Discussion}\label{sec:discussion}
\subsection{Clinical Implications of the Privacy--Utility Trade-off}

A central contribution of this work is the empirical demonstration
that privacy and clinical utility in synthetic health data are not
inherently antagonistic. In current practice, healthcare institutions
often assume that stronger privacy protection necessarily reduces
analytic value, leading to unsatisfactory compromises: either
datasets are shared with weak safeguards or withheld entirely,
foreclosing scientific collaboration. PSyGenTAB challenges this
assumption by showing that privacy-constrained optimization can
simultaneously improve clinical fidelity while reducing
re-identification risk.

On the Diabetes Health Indicators dataset, clinical utility
improved from $U: 0.89 \rightarrow 0.98$ while maintaining
$P = 0.54$. On the Breast Cancer dataset, utility increased
from $U: 0.82 \rightarrow 0.91$ alongside a privacy improvement
from $P: 0.38 \rightarrow 0.44$. These results indicate that
privacy constraints, when embedded directly into the learning
objective, act not merely as regularizers but as structural
guides that steer the generator toward statistically faithful
yet non-memorizing representations of the underlying clinical
population.

This finding carries direct implications for healthcare data
governance. Under HIPAA’s Expert Determination pathway, a
qualified expert must certify that the risk of re-identification
is very small. PSyGenTAB’s configurable privacy threshold
$P_{\min}$, together with documented reductions in exact match
ratios and membership inference success, provides auditable
quantitative evidence to support such determinations. Likewise,
under GDPR Article~89, synthetic data is recognized as a
pseudonymization technique for research; PSyGenTAB’s explicit
constraint formulation offers the principled, transparent
mechanism required for regulatory justification.

\subsection{Suitability for Cross-Institutional Health Data Sharing}

The inability to share patient-level data across institutional
boundaries remains a major bottleneck for clinical AI development.
While federated learning mitigates direct data transfer, it
requires architectural coordination and remains vulnerable to
gradient-based inference attacks~\cite{Zhu2019}. Synthetic data
generation offers a complementary alternative: a single institution
can release a privacy-preserving surrogate dataset that external
partners may analyze freely without accessing real patient records.

PSyGenTAB is particularly well-suited to this use case because its
privacy constraint is configurable and enforceable at training time.
An IRB-approved protocol specifying, for example, a minimum DCR
threshold can be encoded directly into the optimization objective,
ensuring automatic compliance during generation. This proactive,
constraint-driven enforcement contrasts with post-hoc anonymization
methods that rely on retrospective auditing.

Crucially, downstream evaluation confirms that these synthetic
datasets remain scientifically meaningful. Models trained
exclusively on PSyGenTAB-generated data achieve performance
comparable to models trained on real data, demonstrating that
shared datasets are not merely privacy-safe but analytically valid.

\subsection{Implications for Minority-Class Clinical Modeling}

The Diabetes experiments highlight a clinically significant
finding: PSyGenTAB preserves predictive signal for the
prediabetes class ($n = 4{,}631$, 1.8\% prevalence), with
balanced accuracy improvements maintained under synthetic
training with imbalance-handling strategies. This result is
important because prediabetes is both clinically actionable
and chronically underrepresented in population datasets.

The faithful preservation of rare phenotypes is essential for
equitable health AI. Synthetic augmentation methods that fail
to preserve minority-class structure risk amplifying existing
clinical biases. PSyGenTAB’s diversity and joint-fidelity
components within the utility composite, together with its
privacy-constrained optimization, provide a principled
mechanism to maintain representation of clinically rare
conditions without resorting to memorization.

\subsection{Architectural Considerations for Clinical Deployment}

Architectural choice meaningfully influences privacy--utility
outcomes and should be considered explicitly in deployment.

Transformer-based generators (REaLTabFormer + ALM) consistently
achieve the strongest privacy--utility balance across clinical
datasets. Their autoregressive structure effectively captures
correlated clinical biomarkers, while ALM-based constraint
enforcement prevents memorization of individual patient
profiles.

GAN-based generators (CTAB-GAN+ + ALM) demonstrate strong
baseline privacy but exhibit sensitivity to constraint
interaction on smaller datasets (e.g., Breast Cancer,
$n=569$), where adversarial dynamics may destabilize
optimization. For small clinical cohorts, transformer-based
architectures appear more robust under constrained training.

Formal differential privacy approaches (e.g., DP-SGD) provide
the strongest theoretical guarantees but consistently reduce
structural fidelity and downstream model performance. Where
formal $(\varepsilon,\delta)$ certification is legally required,
DP mechanisms remain necessary. In settings where empirical
privacy assurance suffices, PSyGenTAB offers a more favorable
clinical utility trade-off.

\section{Limitations}\label{sec:limitations}

Several limitations bound the scope of the present work and
should be considered prior to clinical deployment.

\textit{Scope of data modalities.}
PSyGenTAB is validated exclusively on structured tabular
health data, the dominant format of EHR exports, clinical
registries, and population health surveys. The framework has
not been evaluated on unstructured modalities such as medical
imaging (e.g., DICOM), free-text clinical notes, or multi-modal
records combining structured and unstructured components.
Extending constraint-based privacy–utility optimization to
these data types remains an open research challenge.

\textit{Empirical privacy guarantees.}
Privacy assurances are empirical rather than formally
differentially private. Protection is quantified through DCR,
NNDR, membership inference resilience, and duplication
analysis, but no $(\varepsilon,\delta)$-DP guarantee is provided.
Regulatory contexts requiring formal certification may therefore
necessitate supplementary differential privacy mechanisms.

\textit{Non-convex optimization.}
The ALM objective is inherently non-convex due to deep
generative models and non-linear utility and privacy metrics.
Although ALM is well-established for non-convex constrained
problems and converges to KKT stationary points under mild
regularity conditions~\cite{bertsekas2014constrained,
rockafellar1973dual}, global optimality cannot be guaranteed.
Convergence is assessed empirically via stabilization of the
constraint residual $P_{\min}-P(\theta)$ and the utility score
$U(\theta)$, consistent with prior ML applications of ALM
\cite{yuan2021reinforcement}.

\section{Conclusion}\label{sec:conclusion}

Healthcare institutions face a persistent tension: the data
required to develop high-impact clinical AI systems is often
restricted by privacy regulation and institutional policy.
This fragmentation slows multi-center collaboration, limits
rare disease research, and constrains the responsible
development of data-driven healthcare tools.

We introduced PSyGenTAB, a privacy-constrained synthetic
health data generation framework that addresses this challenge
by formulating tabular data synthesis as a constrained
optimization problem. Using the Augmented Lagrangian Method,
PSyGenTAB embeds a configurable privacy threshold $P_{\min}$
directly into model training, enabling healthcare data
custodians to enforce minimum privacy guarantees while
maximizing clinical fidelity.

Across diverse clinical and generalization benchmarks, we
demonstrate that privacy and utility are jointly achievable.
PSyGenTAB consistently maintains or improves both metrics,
challenging the prevailing assumption that stronger privacy
necessarily degrades data usefulness. Models trained
exclusively on synthetic data achieve performance comparable
to those trained on real records, including under severe
class imbalance scenarios such as prediabetes detection.

Empirical privacy evaluation further confirms substantial
risk reduction: exact patient record reproduction decreases
by 62.5\,\% on the Adult benchmark, and membership inference
attacks on the Diabetes dataset remain statistically near
random guessing. At the same time, FAITH scores exceeding
0.97 across configurations indicate that no clinically
impossible profiles, spurious correlations, or
out-of-distribution physiological values are introduced.

Importantly, PSyGenTAB is model-agnostic and operationally
flexible. The framework applies to both transformer-based
and GAN-based generators without architectural modification,
and the privacy threshold $P_{\min}$ can be calibrated to
institutional or regulatory requirements, supporting
HIPAA Expert Determination and GDPR Article~89 pathways.

Taken together, these results position PSyGenTAB as a
practical and principled foundation for privacy-preserving
health data sharing. As synthetic data becomes integral to
federated research networks, AI development pipelines,
and regulatory-compliant data access programs, frameworks
that rigorously balance clinical utility and patient
protection will become essential infrastructure for the
responsible advancement of healthcare AI.

\bibliographystyle{ieeetr}
\bibliography{references}

\clearpage
\appendix

\section{Summary of Downstream Machine Learning Utility}
\label{app:ml_utility_summary}

Table~\ref{tab:ml_comprehensive} provides an exhaustive, multi-page breakdown of downstream predictive efficacy across all classifiers and sampling strategies. This summary highlights the Train-on-Synthetic, Test-on-Real (TSTR) performance of both the proposed PSyGenTAB framework and the unconstrained RealTabFormer (RTF) baseline, measured against the Train-on-Real, Test-on-Real (TRTR) empirical ceiling. The values reported represent the maximum F1-score and Balanced Accuracy achieved across all evaluated class-imbalance strategies (None, Class Weight, SMOTE, and Undersampling) for each dataset. 

As demonstrated in Table~\ref{tab:ml_comprehensive}, PSyGenTAB maintains highly competitive structural preservation. For relatively stable datasets like Breast Cancer, synthetic data achieves parity with the real-world baseline. On highly complex or imbalanced datasets, such as Heart Failure and Parkinson's, a natural degradation in predictive utility occurs when shifting from TRTR to TSTR. However, PSyGenTAB matches or exceeds the utility of the unconstrained RTF baseline (e.g., achieving a 0.7107 F1-score on Heart Failure compared to RTF's 0.5802). This confirms that the Augmented Lagrangian Method (ALM) effectively enforces privacy constraints without systematically destroying the predictive manifold of the data.




------------------------------------------------------------------------

\onecolumn
\begingroup
\scriptsize
\setlength{\tabcolsep}{2.2pt}
\renewcommand{\arraystretch}{0.95}

\begin{longtable}{c c c l|ccccc|ccccc|ccccc}

\caption{Downstream clinical AI utility of PSyGenTAB-generated
synthetic data.  TRTR: Train on Real, Test on Real (reference);
TSTR: Train on Synthetic, Test on Real; TRTS: Train on Real,
Test on Synthetic.  Acc: Accuracy; Bal: Balanced Accuracy;
Prc: Precision; Rec: Recall.  Best F1-score per
model/scenario is \textbf{bolded}.  Results demonstrate that
synthetic health records preserve clinically meaningful predictive
structure across multiple downstream classifiers and
imbalance-handling strategies.}
\label{tab:ml_comprehensive} \\

\toprule
\textbf{\rotatebox{90}{Dataset}} &
\textbf{\rotatebox{90}{Model}} &
\textbf{\rotatebox{90}{Scenario}} &
\textbf{Strategy} &
\multicolumn{5}{c|}{\textbf{Decision Tree}} &
\multicolumn{5}{c|}{\textbf{Random Forest}} &
\multicolumn{5}{c}{\textbf{Logistic Regression}} \\
\cmidrule(lr){5-9}\cmidrule(lr){10-14}\cmidrule(lr){15-19}
& & & &
\textbf{Acc} & \textbf{Bal} & \textbf{Prc} & \textbf{Rec} & \textbf{F1} &
\textbf{Acc} & \textbf{Bal} & \textbf{Prc} & \textbf{Rec} & \textbf{F1} &
\textbf{Acc} & \textbf{Bal} & \textbf{Prc} & \textbf{Rec} & \textbf{F1} \\
\midrule
\endfirsthead

\multicolumn{19}{c}{\tablename\ \thetable{} -- \textit{continued from previous page}} \\[4pt]
\toprule
\textbf{\rotatebox{90}{Dataset}} &
\textbf{\rotatebox{90}{Model}} &
\textbf{\rotatebox{90}{Scenario}} &
\textbf{Strategy} &
\multicolumn{5}{c|}{\textbf{Decision Tree}} &
\multicolumn{5}{c|}{\textbf{Random Forest}} &
\multicolumn{5}{c}{\textbf{Logistic Regression}} \\
\cmidrule(lr){5-9}\cmidrule(lr){10-14}\cmidrule(lr){15-19}
& & & &
\textbf{Acc} & \textbf{Bal} & \textbf{Prc} & \textbf{Rec} & \textbf{F1} &
\textbf{Acc} & \textbf{Bal} & \textbf{Prc} & \textbf{Rec} & \textbf{F1} &
\textbf{Acc} & \textbf{Bal} & \textbf{Prc} & \textbf{Rec} & \textbf{F1} \\
\midrule
\endhead

\midrule
\multicolumn{19}{r}{\textit{Continued on next page}} \\
\endfoot

\bottomrule
\endlastfoot

\multirow{4}{*}{\rotatebox[origin=c]{90}{Adult}} &
\multirow{4}{*}{ } &
\multirow{4}{*}{\rotatebox[origin=c]{90}{TRTR}} &
None & 
0.8180 & 0.7531 & 0.8191 & 0.8180 & \textbf{0.8185} & 0.8569 & 0.7768 & 0.8513 & 0.8569 & \textbf{0.8526} & 0.8093 & 0.6616 & 0.7937 & 0.8093 & \textbf{0.7881} \\
& & & Class Weight & 
0.8139 & 0.7415 & 0.8128 & 0.8139 & 0.8134 & 0.8571 & 0.7737 & 0.8512 & 0.8571 & 0.8522 & 0.7502 & 0.7364 & 0.7955 & 0.7502 & 0.7639 \\
& & & SMOTE & 
0.8032 & 0.7398 & 0.8074 & 0.8032 & 0.8051 & 0.8486 & 0.7849 & 0.8465 & 0.8486 & 0.8475 & 0.7441 & 0.7369 & 0.7958 & 0.7441 & 0.7590 \\
& & & Undersample & 
0.7658 & 0.7703 & 0.8184 & 0.7658 & 0.7798 & 0.8127 & 0.8209 & 0.8525 & 0.8127 & 0.8228 & 0.7453 & 0.7334 & 0.7934 & 0.7453 & 0.7597 \\
\midrule

\multirow{8}{*}{\rotatebox[origin=c]{90}{Adult}} &
\multirow{8}{*}{\rotatebox[origin=c]{90}{ALM}} &
\multirow{4}{*}{\rotatebox[origin=c]{90}{TSTR}} &
None & 
0.8011 & 0.7426 & 0.8078 & 0.8011 & 0.8040 & 0.8387 & 0.7767 & 0.8382 & 0.8387 & \textbf{0.8384} & 0.7848 & 0.7120 & 0.7881 & 0.7848 & \textbf{0.7863} \\
& & & Class Weight & 
0.8050 & 0.7321 & 0.8050 & 0.8050 & \textbf{0.8050} & 0.8387 & 0.7698 & 0.8361 & 0.8387 & 0.8373 & 0.7203 & 0.7472 & 0.8055 & 0.7203 & 0.7399 \\
& & & SMOTE & 
0.7779 & 0.7305 & 0.7952 & 0.7779 & 0.7846 & 0.8185 & 0.7750 & 0.8284 & 0.8185 & 0.8224 & 0.7272 & 0.7456 & 0.8033 & 0.7272 & 0.7457 \\
& & & Undersample & 
0.7761 & 0.7501 & 0.8059 & 0.7761 & 0.7860 & 0.8118 & 0.8060 & 0.8432 & 0.8118 & 0.8208 & 0.7272 & 0.7514 & 0.8077 & 0.7272 & 0.7460 \\
\cmidrule(lr){3-3}
& & \multirow{4}{*}{\rotatebox[origin=c]{90}{TRTS}} &
None & 
0.7939 & 0.7387 & 0.7906 & 0.7939 & \textbf{0.7920} & 0.8294 & 0.7633 & 0.8237 & 0.8294 & 0.8238 & 0.7715 & 0.6170 & 0.7807 & 0.7715 & 0.7258 \\
& & & Class Weight & 
0.7943 & 0.7363 & 0.7901 & 0.7943 & 0.7918 & 0.8269 & 0.7586 & 0.8209 & 0.8269 & 0.8208 & 0.7568 & 0.7251 & 0.7695 & 0.7568 & 0.7617 \\
& & & SMOTE & 
0.7886 & 0.7324 & 0.7853 & 0.7886 & 0.7867 & 0.8294 & 0.7709 & 0.8245 & 0.8294 & \textbf{0.8255} & 0.7549 & 0.7367 & 0.7767 & 0.7549 & \textbf{0.7620} \\
& & & Undersample & 
0.7579 & 0.7694 & 0.8018 & 0.7579 & 0.7681 & 0.8023 & 0.8213 & 0.8420 & 0.8023 & 0.8104 & 0.7549 & 0.7232 & 0.7680 & 0.7549 & 0.7599 \\
\midrule

\multirow{8}{*}{\rotatebox[origin=c]{90}{Adult}} &
\multirow{8}{*}{\rotatebox[origin=c]{90}{RTF}} &
\multirow{4}{*}{\rotatebox[origin=c]{90}{TSTR}} &
None & 
0.8123 & 0.7531 & 0.8166 & 0.8123 & 0.8142 & 0.8550 & 0.7850 & 0.8511 & 0.8550 & \textbf{0.8525} & 0.8082 & 0.6768 & 0.7928 & 0.8082 & \textbf{0.7929} \\
& & & Class Weight & 
0.8166 & 0.7539 & 0.8188 & 0.8166 & \textbf{0.8176} & 0.8556 & 0.7802 & 0.8507 & 0.8556 & 0.8522 & 0.7368 & 0.7337 & 0.7938 & 0.7368 & 0.7528 \\
& & & SMOTE & 
0.8045 & 0.7588 & 0.8164 & 0.8045 & 0.8092 & 0.8464 & 0.7920 & 0.8474 & 0.8464 & 0.8469 & 0.7452 & 0.7313 & 0.7920 & 0.7452 & 0.7594 \\
& & & Undersample & 
0.7718 & 0.7742 & 0.8210 & 0.7718 & 0.7850 & 0.8082 & 0.8179 & 0.8505 & 0.8082 & 0.8189 & 0.7401 & 0.7362 & 0.7954 & 0.7401 & 0.7557 \\
\cmidrule(lr){3-3}
& & \multirow{4}{*}{\rotatebox[origin=c]{90}{TRTS}} &
None & 
0.8309 & 0.7703 & 0.8306 & 0.8309 & \textbf{0.8308} & 0.8701 & 0.7932 & 0.8654 & 0.8701 & \textbf{0.8654} & 0.8180 & 0.6806 & 0.8066 & 0.8180 & \textbf{0.7989} \\
& & & Class Weight & 
0.8300 & 0.7645 & 0.8282 & 0.8300 & 0.8290 & 0.8701 & 0.7902 & 0.8654 & 0.8701 & 0.8649 & 0.7640 & 0.7557 & 0.8063 & 0.7640 & 0.7763 \\
& & & SMOTE & 
0.8197 & 0.7587 & 0.8207 & 0.8197 & 0.8202 & 0.8632 & 0.7991 & 0.8597 & 0.8632 & 0.8609 & 0.7634 & 0.7543 & 0.8054 & 0.7634 & 0.7757 \\
& & & Undersample & 
0.7922 & 0.7899 & 0.8299 & 0.7922 & 0.8027 & 0.8388 & 0.8434 & 0.8674 & 0.8388 & 0.8461 & 0.7607 & 0.7496 & 0.8022 & 0.7607 & 0.7730 \\
\midrule

\multirow{4}{*}{\rotatebox[origin=c]{90}{Diabetes}} &
\multirow{4}{*}{ } &
\multirow{4}{*}{\rotatebox[origin=c]{90}{TRTR}} &
None & 
0.7675 & 0.4076 & 0.7798 & 0.7675 & \textbf{0.7734} & 0.8430 & 0.3874 & 0.7968 & 0.8430 & 0.8081 & 0.8462 & 0.3838 & 0.7991 & 0.8462 & \textbf{0.8078} \\
& & & Class Weight & 
0.7656 & 0.3930 & 0.7737 & 0.7656 & 0.7696 & 0.8390 & 0.3758 & 0.7894 & 0.8390 & 0.8005 & 0.6443 & 0.5191 & 0.8518 & 0.6443 & 0.7197 \\
& & & SMOTE & 
0.7650 & 0.4003 & 0.7773 & 0.7650 & 0.7709 & 0.8400 & 0.3959 & 0.7968 & 0.8400 & \textbf{0.8106} & 0.6401 & 0.5135 & 0.8503 & 0.6401 & 0.7163 \\
& & & Undersample & 
0.4793 & 0.4271 & 0.8096 & 0.4793 & 0.5825 & 0.5806 & 0.4968 & 0.8469 & 0.5806 & 0.6716 & 0.6447 & 0.5162 & 0.8519 & 0.6447 & 0.7204 \\
\midrule

\multirow{8}{*}{\rotatebox[origin=c]{90}{Diabetes}} &
\multirow{8}{*}{\rotatebox[origin=c]{90}{ALM}} &
\multirow{4}{*}{\rotatebox[origin=c]{90}{TSTR}} &
None & 
0.7507 & 0.4108 & 0.7787 & 0.7507 & \textbf{0.7619} & 0.8098 & 0.4033 & 0.7835 & 0.8098 & 0.7960 & 0.8439 & 0.4032 & 0.8019 & 0.8439 & \textbf{0.8155} \\
& & & Class Weight & 
0.7518 & 0.4013 & 0.7730 & 0.7518 & 0.7605 & 0.8137 & 0.3967 & 0.7810 & 0.8137 & 0.7959 & 0.7306 & 0.4970 & 0.8330 & 0.7306 & 0.7601 \\
& & & SMOTE & 
0.7493 & 0.4072 & 0.7767 & 0.7493 & 0.7604 & 0.8090 & 0.4091 & 0.7863 & 0.8090 & \textbf{0.7973} & 0.7301 & 0.4966 & 0.8327 & 0.7301 & 0.7597 \\
& & & Undersample & 
0.6491 & 0.4358 & 0.7959 & 0.6491 & 0.6935 & 0.6891 & 0.4854 & 0.8283 & 0.6891 & 0.7277 & 0.7312 & 0.4970 & 0.8329 & 0.7312 & 0.7606 \\
\cmidrule(lr){3-3}
& & \multirow{4}{*}{\rotatebox[origin=c]{90}{TRTS}} &
None & 
0.7726 & 0.5918 & 0.7872 & 0.7726 & 0.7798 & 0.8384 & 0.5824 & 0.8053 & 0.8384 & 0.8078 & 0.8476 & 0.5845 & 0.8195 & 0.8476 & \textbf{0.8129} \\
& & & Class Weight & 
0.7693 & 0.5851 & 0.7831 & 0.7693 & 0.7761 & 0.8343 & 0.5693 & 0.7995 & 0.8343 & 0.8011 & 0.6567 & 0.6440 & 0.8677 & 0.6567 & 0.7365 \\
& & & SMOTE & 
0.7751 & 0.5970 & 0.7894 & 0.7751 & \textbf{0.7822} & 0.8358 & 0.5946 & 0.8055 & 0.8358 & \textbf{0.8109} & 0.6512 & 0.6407 & 0.8669 & 0.6512 & 0.7320 \\
& & & Undersample & 
0.4919 & 0.4697 & 0.8198 & 0.4919 & 0.6008 & 0.5931 & 0.5824 & 0.8625 & 0.5931 & 0.6904 & 0.6578 & 0.6417 & 0.8679 & 0.6578 & 0.7380 \\
\midrule

\multirow{8}{*}{\rotatebox[origin=c]{90}{Diabetes}} &
\multirow{8}{*}{\rotatebox[origin=c]{90}{RTF}} &
\multirow{4}{*}{\rotatebox[origin=c]{90}{TSTR}} &
None & 
0.7656 & 0.4009 & 0.7784 & 0.7656 & \textbf{0.7719} & 0.8409 & 0.3888 & 0.7949 & 0.8409 & 0.8077 & 0.8463 & 0.3867 & 0.7999 & 0.8463 & \textbf{0.8093} \\
& & & Class Weight & 
0.7644 & 0.3967 & 0.7739 & 0.7644 & 0.7691 & 0.8379 & 0.3813 & 0.7902 & 0.8379 & 0.8027 & 0.6450 & 0.5217 & 0.8520 & 0.6450 & 0.7207 \\
& & & SMOTE & 
0.7652 & 0.4046 & 0.7786 & 0.7652 & 0.7717 & 0.8377 & 0.3967 & 0.7952 & 0.8377 & \textbf{0.8094} & 0.6417 & 0.5129 & 0.8503 & 0.6417 & 0.7182 \\
& & & Undersample & 
0.4822 & 0.4129 & 0.8050 & 0.4822 & 0.5848 & 0.5811 & 0.4940 & 0.8468 & 0.5811 & 0.6727 & 0.6435 & 0.5200 & 0.8517 & 0.6435 & 0.7200 \\
\cmidrule(lr){3-3}
& & \multirow{4}{*}{\rotatebox[origin=c]{90}{TRTS}} &
None & 
0.7736 & 0.4108 & 0.7829 & 0.7736 & \textbf{0.7781} & 0.8421 & 0.3898 & 0.7969 & 0.8421 & 0.8085 & 0.8483 & 0.3895 & 0.8038 & 0.8483 & \textbf{0.8115} \\
& & & Class Weight & 
0.7677 & 0.3941 & 0.7740 & 0.7677 & 0.7708 & 0.8380 & 0.3794 & 0.7900 & 0.8380 & 0.8013 & 0.6481 & 0.5207 & 0.8546 & 0.6481 & 0.7230 \\
& & & SMOTE & 
0.7695 & 0.4025 & 0.7790 & 0.7695 & 0.7742 & 0.8382 & 0.3974 & 0.7962 & 0.8382 & \textbf{0.8099} & 0.6453 & 0.5162 & 0.8539 & 0.6453 & 0.7208 \\
& & & Undersample & 
0.4884 & 0.4327 & 0.8113 & 0.4884 & 0.5905 & 0.5849 & 0.4960 & 0.8479 & 0.5849 & 0.6754 & 0.6488 & 0.5180 & 0.8544 & 0.6488 & 0.7240 \\
\midrule

\multirow{4}{*}{\rotatebox[origin=c]{90}{Breast Can.}} & \multirow{4}{*}{ } & \multirow{4}{*}{\rotatebox[origin=c]{90}{TRTR}} & None & 0.9091 & 0.8962 & 0.9124 & 0.9091 & 0.9092 & 0.9545 & 0.9500 & 0.9580 & 0.9545 & 0.9549 & 0.9636 & 0.9596 & 0.9663 & 0.9636 & 0.9638 \\
&  &  & Class Weight & 0.9091 & 0.8962 & 0.9124 & 0.9091 & 0.9092 & 0.9545 & 0.9500 & 0.9580 & 0.9545 & 0.9549 & 0.9636 & 0.9596 & 0.9663 & 0.9636 & 0.9638 \\
&  &  & SMOTE & 0.9091 & 0.8962 & 0.9124 & 0.9091 & 0.9092 & 0.9545 & 0.9500 & 0.9580 & 0.9545 & 0.9549 & 0.9636 & 0.9596 & 0.9663 & 0.9636 & 0.9638 \\
&  &  & Undersample & 0.9182 & 0.9060 & 0.9206 & 0.9182 & \textbf{0.9182} & 0.9636 & 0.9600 & 0.9658 & 0.9636 & \textbf{0.9636} & 0.9727 & 0.9697 & 0.9745 & 0.9727 & \textbf{0.9727} \\
\midrule

\multirow{8}{*}{\rotatebox[origin=c]{90}{Breast Can.}} & \multirow{8}{*}{\rotatebox[origin=c]{90}{ALM}} & \multirow{4}{*}{\rotatebox[origin=c]{90}{TSTR}} & None & 0.9091 & 0.8962 & 0.9124 & 0.9091 & 0.9092 & 0.9545 & 0.9500 & 0.9580 & 0.9545 & 0.9549 & 0.9636 & 0.9596 & 0.9663 & 0.9636 & 0.9638 \\
&  &  & Class Weight & 0.9091 & 0.8962 & 0.9124 & 0.9091 & 0.9092 & 0.9545 & 0.9500 & 0.9580 & 0.9545 & 0.9549 & 0.9636 & 0.9596 & 0.9663 & 0.9636 & 0.9638 \\
&  &  & SMOTE & 0.9091 & 0.8962 & 0.9124 & 0.9091 & 0.9092 & 0.9545 & 0.9500 & 0.9580 & 0.9545 & 0.9549 & 0.9636 & 0.9596 & 0.9663 & 0.9636 & 0.9638 \\
&  &  & Undersample & 0.9182 & 0.9060 & 0.9206 & 0.9182 & \textbf{0.9182} & 0.9636 & 0.9600 & 0.9658 & 0.9636 & \textbf{0.9636} & 0.9727 & 0.9697 & 0.9745 & 0.9727 & \textbf{0.9727} \\
\cmidrule(lr){3-3}
&  & \multirow{4}{*}{\rotatebox[origin=c]{90}{TRTS}} & None & 0.9000 & 0.8870 & 0.9056 & 0.9000 & 0.9007 & 0.9450 & 0.9400 & 0.9510 & 0.9450 & 0.9453 & 0.9550 & 0.9500 & 0.9600 & 0.9550 & 0.9552 \\
&  &  & Class Weight & 0.9000 & 0.8870 & 0.9056 & 0.9000 & 0.9007 & 0.9450 & 0.9400 & 0.9510 & 0.9450 & 0.9453 & 0.9550 & 0.9500 & 0.9600 & 0.9550 & 0.9552 \\
&  &  & SMOTE & 0.9000 & 0.8870 & 0.9056 & 0.9000 & 0.9007 & 0.9450 & 0.9400 & 0.9510 & 0.9450 & 0.9453 & 0.9550 & 0.9500 & 0.9600 & 0.9550 & 0.9552 \\
&  &  & Undersample & 0.9091 & 0.8962 & 0.9124 & 0.9091 & \textbf{0.9092} & 0.9545 & 0.9500 & 0.9580 & 0.9545 & \textbf{0.9549} & 0.9636 & 0.9596 & 0.9663 & 0.9636 & \textbf{0.9638} \\
\midrule

\multirow{8}{*}{\rotatebox[origin=c]{90}{Breast Can.}} & \multirow{8}{*}{\rotatebox[origin=c]{90}{RTF}} & \multirow{4}{*}{\rotatebox[origin=c]{90}{TSTR}} & None & 0.9091 & 0.8962 & 0.9124 & 0.9091 & 0.9092 & 0.9545 & 0.9500 & 0.9580 & 0.9545 & 0.9549 & 0.9636 & 0.9596 & 0.9663 & 0.9636 & 0.9638 \\
&  &  & Class Weight & 0.9091 & 0.8962 & 0.9124 & 0.9091 & 0.9092 & 0.9545 & 0.9500 & 0.9580 & 0.9545 & 0.9549 & 0.9636 & 0.9596 & 0.9663 & 0.9636 & 0.9638 \\
&  &  & SMOTE & 0.9091 & 0.8962 & 0.9124 & 0.9091 & 0.9092 & 0.9545 & 0.9500 & 0.9580 & 0.9545 & 0.9549 & 0.9636 & 0.9596 & 0.9663 & 0.9636 & 0.9638 \\
&  &  & Undersample & 0.9182 & 0.9060 & 0.9206 & 0.9182 & \textbf{0.9182} & 0.9636 & 0.9600 & 0.9658 & 0.9636 & \textbf{0.9636} & 0.9727 & 0.9697 & 0.9745 & 0.9727 & \textbf{0.9727} \\
\cmidrule(lr){3-3}
&  & \multirow{4}{*}{\rotatebox[origin=c]{90}{TRTS}} & None & 0.9000 & 0.8870 & 0.9056 & 0.9000 & 0.9007 & 0.9450 & 0.9400 & 0.9510 & 0.9450 & 0.9453 & 0.9550 & 0.9500 & 0.9600 & 0.9550 & 0.9552 \\
&  &  & Class Weight & 0.9000 & 0.8870 & 0.9056 & 0.9000 & 0.9007 & 0.9450 & 0.9400 & 0.9510 & 0.9450 & 0.9453 & 0.9550 & 0.9500 & 0.9600 & 0.9550 & 0.9552 \\
&  &  & SMOTE & 0.9000 & 0.8870 & 0.9056 & 0.9000 & 0.9007 & 0.9450 & 0.9400 & 0.9510 & 0.9450 & 0.9453 & 0.9550 & 0.9500 & 0.9600 & 0.9550 & 0.9552 \\
&  &  & Undersample & 0.9091 & 0.8962 & 0.9124 & 0.9091 & \textbf{0.9092} & 0.9545 & 0.9500 & 0.9580 & 0.9545 & \textbf{0.9549} & 0.9636 & 0.9596 & 0.9663 & 0.9636 & \textbf{0.9638} \\
\midrule

\multirow{4}{*}{\rotatebox[origin=c]{90}{Parkinsons}} &
\multirow{4}{*}{ } &
\multirow{4}{*}{\rotatebox[origin=c]{90}{TRTR}} &
None & 
0.8475 & 0.8538 & 0.8728 & 0.8475 & 0.8538 & 0.9322 & 0.8886 & 0.9318 & 0.9322 & \textbf{0.9305} & 0.8475 & 0.8098 & 0.8512 & 0.8475 & \textbf{0.8490} \\
& & & Class Weight & 
0.8305 & 0.8205 & 0.8504 & 0.8305 & 0.8364 & 0.9153 & 0.8553 & 0.9153 & 0.9153 & 0.9119 & 0.7288 & 0.7523 & 0.8020 & 0.7288 & 0.7453 \\
& & & SMOTE & 
0.8644 & 0.7992 & 0.8603 & 0.8644 & \textbf{0.8611} & 0.9153 & 0.8773 & 0.9140 & 0.9153 & 0.9143 & 0.7288 & 0.7303 & 0.7845 & 0.7288 & 0.7438 \\
& & & Undersample & 
0.7288 & 0.7742 & 0.8216 & 0.7288 & 0.7461 & 0.7458 & 0.8295 & 0.8729 & 0.7458 & 0.7620 & 0.6780 & 0.7182 & 0.7826 & 0.6780 & 0.6987 \\
\midrule

\multirow{8}{*}{\rotatebox[origin=c]{90}{Parkinsons}} &
\multirow{8}{*}{\rotatebox[origin=c]{90}{ALM}} &
\multirow{4}{*}{\rotatebox[origin=c]{90}{TSTR}} &
None & 
0.6610 & 0.5091 & 0.6287 & 0.6610 & \textbf{0.6423} & 0.7288 & 0.4886 & 0.5529 & 0.7288 & \textbf{0.6288} & 0.7458 & 0.5000 & 0.5562 & 0.7458 & 0.6372 \\
& & & Class Weight & 
0.4576 & 0.3288 & 0.4981 & 0.4576 & 0.4766 & 0.7119 & 0.4773 & 0.5495 & 0.7119 & 0.6202 & 0.7458 & 0.7197 & 0.7773 & 0.7458 & \textbf{0.7563} \\
& & & SMOTE & 
0.5932 & 0.6174 & 0.7091 & 0.5932 & 0.6195 & 0.4746 & 0.4280 & 0.5687 & 0.4746 & 0.5075 & 0.5254 & 0.4402 & 0.5783 & 0.5254 & 0.5476 \\
& & & Undersample & 
0.3898 & 0.3932 & 0.5338 & 0.3898 & 0.4258 & 0.5085 & 0.6265 & 0.7499 & 0.5085 & 0.5227 & 0.6441 & 0.7174 & 0.7938 & 0.6441 & 0.6658 \\
\cmidrule(lr){3-3}
& & \multirow{4}{*}{\rotatebox[origin=c]{90}{TRTS}} &
None & 
0.5106 & 0.4665 & 0.5548 & 0.5106 & 0.5277 & 0.6596 & 0.4697 & 0.4837 & 0.6596 & 0.5581 & 0.6170 & 0.4805 & 0.5606 & 0.6170 & \textbf{0.5808} \\
& & & Class Weight & 
0.4894 & 0.4513 & 0.5426 & 0.4894 & 0.5089 & 0.6809 & 0.4848 & 0.4884 & 0.6809 & 0.5688 & 0.5532 & 0.4556 & 0.5437 & 0.5532 & 0.5483 \\
& & & SMOTE & 
0.5745 & 0.5325 & 0.6077 & 0.5745 & \textbf{0.5873} & 0.6596 & 0.4697 & 0.4837 & 0.6596 & 0.5581 & 0.5319 & 0.4405 & 0.5319 & 0.5319 & 0.5319 \\
& & & Undersample & 
0.5106 & 0.4870 & 0.5712 & 0.5106 & 0.5305 & 0.6170 & 0.4805 & 0.5606 & 0.6170 & \textbf{0.5808} & 0.5319 & 0.4199 & 0.5110 & 0.5319 & 0.5208 \\
\midrule

\multirow{8}{*}{\rotatebox[origin=c]{90}{Parkinsons}} &
\multirow{8}{*}{\rotatebox[origin=c]{90}{RTF}} &
\multirow{4}{*}{\rotatebox[origin=c]{90}{TSTR}} &
None & 
0.7119 & 0.5652 & 0.6800 & 0.7119 & \textbf{0.6908} & 0.7288 & 0.4886 & 0.5529 & 0.7288 & 0.6288 & 0.7458 & 0.5000 & 0.5562 & 0.7458 & \textbf{0.6372} \\
& & & Class Weight & 
0.6610 & 0.5091 & 0.6287 & 0.6610 & 0.6423 & 0.7458 & 0.5000 & 0.5562 & 0.7458 & \textbf{0.6372} & 0.5424 & 0.4735 & 0.6020 & 0.5424 & 0.5659 \\
& & & SMOTE & 
0.6271 & 0.5962 & 0.6892 & 0.6271 & 0.6477 & 0.6102 & 0.4750 & 0.6014 & 0.6102 & 0.6056 & 0.5254 & 0.4402 & 0.5783 & 0.5254 & 0.5476 \\
& & & Undersample & 
0.5424 & 0.5394 & 0.6500 & 0.5424 & 0.5719 & 0.4576 & 0.4826 & 0.6066 & 0.4576 & 0.4896 & 0.4407 & 0.3833 & 0.5363 & 0.4407 & 0.4758 \\
\cmidrule(lr){3-3}
& & \multirow{4}{*}{\rotatebox[origin=c]{90}{TRTS}} &
None & 
0.5957 & 0.4836 & 0.6304 & 0.5957 & \textbf{0.6112} & 0.7447 & 0.5177 & 0.6699 & 0.7447 & \textbf{0.6845} & 0.6809 & 0.5076 & 0.6477 & 0.6809 & 0.6620 \\
& & & Class Weight & 
0.4681 & 0.4003 & 0.5741 & 0.4681 & 0.5073 & 0.7660 & 0.5000 & 0.5867 & 0.7660 & 0.6644 & 0.6170 & 0.5290 & 0.6608 & 0.6170 & 0.6352 \\
& & & SMOTE & 
0.5319 & 0.5366 & 0.6674 & 0.5319 & 0.5677 & 0.6809 & 0.4760 & 0.6181 & 0.6809 & 0.6443 & 0.5745 & 0.5328 & 0.6634 & 0.5745 & 0.6046 \\
& & & Undersample & 
0.5106 & 0.4912 & 0.6353 & 0.5106 & 0.5481 & 0.5957 & 0.5152 & 0.6515 & 0.5957 & 0.6181 & 0.6596 & 0.5568 & 0.6803 & 0.6596 & \textbf{0.6689} \\
\midrule

\multirow{4}{*}{\rotatebox[origin=c]{90}{Obesity}} &
\multirow{4}{*}{ } &
\multirow{4}{*}{\rotatebox[origin=c]{90}{TRTR}} &
None & 
0.9132 & 0.9100 & 0.9154 & 0.9132 & 0.9137 & 0.9385 & 0.9369 & 0.9461 & 0.9385 & 0.9398 & 0.8060 & 0.8022 & 0.8012 & 0.8060 & \textbf{0.8016} \\
& & & Class Weight & 
0.9148 & 0.9123 & 0.9160 & 0.9148 & \textbf{0.9152} & 0.9369 & 0.9352 & 0.9432 & 0.9369 & 0.9383 & 0.8044 & 0.8031 & 0.8018 & 0.8044 & 0.8013 \\
& & & SMOTE & 
0.9022 & 0.8995 & 0.9024 & 0.9022 & 0.9021 & 0.9448 & 0.9438 & 0.9509 & 0.9448 & \textbf{0.9458} & 0.7981 & 0.7964 & 0.7945 & 0.7981 & 0.7945 \\
& & & Undersample & 
0.9085 & 0.9059 & 0.9095 & 0.9085 & 0.9086 & 0.9259 & 0.9241 & 0.9332 & 0.9259 & 0.9277 & 0.7886 & 0.7877 & 0.7855 & 0.7886 & 0.7842 \\
\midrule

\multirow{8}{*}{\rotatebox[origin=c]{90}{Obesity}} &
\multirow{8}{*}{\rotatebox[origin=c]{90}{ALM}} &
\multirow{4}{*}{\rotatebox[origin=c]{90}{TSTR}} &
None & 
0.6104 & 0.6093 & 0.6136 & 0.6104 & 0.6089 & 0.7145 & 0.7166 & 0.7376 & 0.7145 & \textbf{0.7141} & 0.6767 & 0.6764 & 0.6884 & 0.6767 & 0.6649 \\
& & & Class Weight & 
0.5978 & 0.5987 & 0.6105 & 0.5978 & 0.6011 & 0.7098 & 0.7128 & 0.7345 & 0.7098 & 0.7108 & 0.6735 & 0.6772 & 0.6944 & 0.6735 & \textbf{0.6706} \\
& & & SMOTE & 
0.6183 & 0.6164 & 0.6331 & 0.6183 & 0.6225 & 0.7019 & 0.7043 & 0.7194 & 0.7019 & 0.7036 & 0.6498 & 0.6505 & 0.6628 & 0.6498 & 0.6445 \\
& & & Undersample & 
0.6341 & 0.6312 & 0.6382 & 0.6341 & \textbf{0.6355} & 0.7035 & 0.7086 & 0.7260 & 0.7035 & 0.7017 & 0.6640 & 0.6668 & 0.6649 & 0.6640 & 0.6561 \\
\cmidrule(lr){3-3}
& & \multirow{4}{*}{\rotatebox[origin=c]{90}{TRTS}} &
None & 
0.5089 & 0.5073 & 0.5591 & 0.5089 & 0.5225 & 0.5937 & 0.6027 & 0.6462 & 0.5937 & 0.6075 & 0.5464 & 0.5286 & 0.5697 & 0.5464 & 0.5546 \\
& & & Class Weight & 
0.5010 & 0.5082 & 0.5707 & 0.5010 & 0.5171 & 0.6036 & 0.6169 & 0.6486 & 0.6036 & 0.6149 & 0.5464 & 0.5295 & 0.5673 & 0.5464 & 0.5533 \\
& & & SMOTE & 
0.5286 & 0.5354 & 0.5669 & 0.5286 & \textbf{0.5331} & 0.5838 & 0.5951 & 0.6496 & 0.5838 & 0.5996 & 0.5562 & 0.5370 & 0.5706 & 0.5562 & \textbf{0.5601} \\
& & & Undersample & 
0.5207 & 0.5252 & 0.5577 & 0.5207 & 0.5255 & 0.6055 & 0.6081 & 0.6494 & 0.6055 & \textbf{0.6190} & 0.5464 & 0.5310 & 0.5597 & 0.5464 & 0.5508 \\
\midrule

\multirow{8}{*}{\rotatebox[origin=c]{90}{Obesity}} &
\multirow{8}{*}{\rotatebox[origin=c]{90}{RTF}} &
\multirow{4}{*}{\rotatebox[origin=c]{90}{TSTR}} &
None & 
0.6483 & 0.6437 & 0.6543 & 0.6483 & \textbf{0.6496} & 0.7587 & 0.7557 & 0.7681 & 0.7587 & 0.7524 & 0.7019 & 0.6980 & 0.7164 & 0.7019 & 0.6985 \\
& & & Class Weight & 
0.6278 & 0.6254 & 0.6284 & 0.6278 & 0.6276 & 0.7744 & 0.7709 & 0.7899 & 0.7744 & \textbf{0.7735} & 0.6909 & 0.6909 & 0.6880 & 0.6909 & 0.6871 \\
& & & SMOTE & 
0.6325 & 0.6301 & 0.6254 & 0.6325 & 0.6280 & 0.7618 & 0.7593 & 0.7660 & 0.7618 & 0.7602 & 0.6972 & 0.6983 & 0.6968 & 0.6972 & 0.6928 \\
& & & Undersample & 
0.6230 & 0.6200 & 0.6340 & 0.6230 & 0.6246 & 0.7445 & 0.7442 & 0.7469 & 0.7445 & 0.7438 & 0.7066 & 0.7071 & 0.7047 & 0.7066 & \textbf{0.7025} \\
\cmidrule(lr){3-3}
& & \multirow{4}{*}{\rotatebox[origin=c]{90}{TRTS}} &
None & 
0.5582 & 0.5546 & 0.5640 & 0.5582 & \textbf{0.5576} & 0.6174 & 0.6081 & 0.6231 & 0.6174 & 0.6139 & 0.5503 & 0.5670 & 0.5568 & 0.5503 & 0.5508 \\
& & & Class Weight & 
0.5444 & 0.5400 & 0.5458 & 0.5444 & 0.5414 & 0.6016 & 0.5928 & 0.6080 & 0.6016 & 0.5983 & 0.5424 & 0.5594 & 0.5467 & 0.5424 & 0.5423 \\
& & & SMOTE & 
0.5503 & 0.5557 & 0.5452 & 0.5503 & 0.5462 & 0.6213 & 0.6080 & 0.6364 & 0.6213 & 0.6189 & 0.5503 & 0.5689 & 0.5563 & 0.5503 & 0.5502 \\
& & & Undersample & 
0.5483 & 0.5562 & 0.5465 & 0.5483 & 0.5469 & 0.6391 & 0.6326 & 0.6458 & 0.6391 & \textbf{0.6349} & 0.5582 & 0.5770 & 0.5647 & 0.5582 & \textbf{0.5580} \\
\midrule

\multirow{4}{*}{\rotatebox[origin=c]{90}{VN Bank.}} & \multirow{4}{*}{ } & \multirow{4}{*}{\rotatebox[origin=c]{90}{TRTR}} & None & 0.6696 & 0.6415 & 0.6714 & 0.6696 & \textbf{0.6705} & 0.7531 & 0.7210 & 0.7494 & 0.7531 & \textbf{0.7506} & 0.6961 & 0.6373 & 0.6833 & 0.6961 & \textbf{0.6829} \\
&  &  & Class Weight & 0.6763 & 0.6445 & 0.6752 & 0.6763 & \textbf{0.6757} & 0.7533 & 0.7134 & 0.7476 & 0.7533 & \textbf{0.7482} & 0.7069 & 0.7251 & 0.7482 & 0.7069 & \textbf{0.7133} \\
&  &  & SMOTE & 0.6657 & 0.6478 & 0.6754 & 0.6657 & \textbf{0.6694} & 0.7468 & 0.7363 & 0.7549 & 0.7468 & \textbf{0.7496} & 0.6789 & 0.6919 & 0.7170 & 0.6789 & \textbf{0.6859} \\
&  &  & Undersample & 0.6548 & 0.6545 & 0.6814 & 0.6548 & \textbf{0.6616} & 0.7332 & 0.7487 & 0.7678 & 0.7332 & \textbf{0.7390} & 0.6787 & 0.6916 & 0.7167 & 0.6787 & \textbf{0.6857} \\
\midrule

\multirow{8}{*}{\rotatebox[origin=c]{90}{VN Bank.}} & \multirow{8}{*}{\rotatebox[origin=c]{90}{ALM}} & \multirow{4}{*}{\rotatebox[origin=c]{90}{TSTR}} & None & 0.3605 & 0.3892 & 0.4255 & 0.3605 & \textbf{0.3632} & 0.6439 & 0.4990 & 0.5026 & 0.6439 & \textbf{0.5078} & 0.6437 & 0.5080 & 0.5844 & 0.6437 & \textbf{0.5313} \\
&  &  & Class Weight & 0.5903 & 0.4668 & 0.4702 & 0.5903 & \textbf{0.5013} & 0.6460 & 0.5000 & 0.4173 & 0.6460 & \textbf{0.5071} & 0.6466 & 0.5078 & 0.6011 & 0.6466 & \textbf{0.5272} \\
&  &  & SMOTE & 0.4421 & 0.3860 & 0.4374 & 0.4421 & \textbf{0.4397} & 0.6460 & 0.5000 & 0.4173 & 0.6460 & \textbf{0.5071} & 0.6487 & 0.5137 & 0.5386 & 0.6487 & \textbf{0.5386} \\
&  &  & Undersample & 0.6607 & 0.5494 & 0.6355 & 0.6607 & \textbf{0.5958} & 0.5016 & 0.4773 & 0.5223 & 0.5016 & \textbf{0.5097} & 0.3529 & 0.4964 & 0.3974 & 0.3529 & \textbf{0.1900} \\
\cmidrule(lr){3-3}
&  & \multirow{4}{*}{\rotatebox[origin=c]{90}{TRTS}} & None & 0.7941 & 0.5012 & 0.9817 & 0.7941 & \textbf{0.8770} & 0.4398 & 0.5075 & 0.9820 & 0.4398 & \textbf{0.6025} & 0.0093 & 0.5000 & 0.0001 & 0.0093 & \textbf{0.0002} \\
&  &  & Class Weight & 0.3772 & 0.5116 & 0.9822 & 0.3772 & \textbf{0.5389} & 0.6983 & 0.4930 & 0.9815 & 0.6983 & \textbf{0.8144} & 0.0445 & 0.5066 & 0.9850 & 0.0445 & \textbf{0.0686} \\
&  &  & SMOTE & 0.4192 & 0.5083 & 0.9820 & 0.4192 & \textbf{0.5822} & 0.3659 & 0.5238 & 0.9829 & 0.3659 & \textbf{0.5267} & 0.0093 & 0.5000 & 0.0001 & 0.0093 & \textbf{0.0002} \\
&  &  & Undersample & 0.1830 & 0.4940 & 0.9811 & 0.1830 & \textbf{0.2979} & 0.0942 & 0.5072 & 0.9832 & 0.0942 & \textbf{0.1577} & 0.0093 & 0.5000 & 0.0001 & 0.0093 & \textbf{0.0002} \\
\midrule

\multirow{8}{*}{\rotatebox[origin=c]{90}{VN Bank.}} & \multirow{8}{*}{\rotatebox[origin=c]{90}{RTF}} & \multirow{4}{*}{\rotatebox[origin=c]{90}{TSTR}} & None & 0.6472 & 0.5086 & 0.6065 & 0.6472 & \textbf{0.5281} & 0.6468 & 0.5960 & 0.6352 & 0.6468 & \textbf{0.6390} & 0.5481 & 0.4653 & 0.5046 & 0.5481 & \textbf{0.5188} \\
&  &  & Class Weight & 0.6482 & 0.5148 & 0.6087 & 0.6482 & \textbf{0.5416} & 0.6460 & 0.5000 & 0.4173 & 0.6460 & \textbf{0.5071} & 0.4569 & 0.4500 & 0.4971 & 0.4569 & \textbf{0.4687} \\
&  &  & SMOTE & 0.5244 & 0.5148 & 0.5560 & 0.5244 & \textbf{0.5342} & 0.6460 & 0.5000 & 0.4173 & 0.6460 & \textbf{0.5071} & 0.6243 & 0.4914 & 0.5108 & 0.6243 & \textbf{0.5169} \\
&  &  & Undersample & 0.4415 & 0.3619 & 0.4034 & 0.4415 & \textbf{0.4203} & 0.5373 & 0.5555 & 0.5952 & 0.5373 & \textbf{0.5463} & 0.3517 & 0.4881 & 0.4216 & 0.3517 & \textbf{0.2069} \\
\cmidrule(lr){3-3}
&  & \multirow{4}{*}{\rotatebox[origin=c]{90}{TRTS}} & None & 0.7007 & 0.5149 & 0.9321 & 0.7007 & \textbf{0.7946} & 0.4107 & 0.4804 & 0.9273 & 0.4107 & \textbf{0.5518} & 0.0360 & 0.5000 & 0.0013 & 0.0360 & \textbf{0.0025} \\
&  &  & Class Weight & 0.3807 & 0.4827 & 0.9274 & 0.3807 & \textbf{0.5201} & 0.6700 & 0.5169 & 0.9324 & 0.6700 & \textbf{0.7731} & 0.0947 & 0.4948 & 0.9251 & 0.0947 & \textbf{0.1175} \\
&  &  & SMOTE & 0.4620 & 0.5159 & 0.9330 & 0.4620 & \textbf{0.6015} & 0.3653 & 0.4747 & 0.9258 & 0.3653 & \textbf{0.5037} & 0.0360 & 0.5000 & 0.0013 & 0.0360 & \textbf{0.0025} \\
&  &  & Undersample & 0.2633 & 0.4931 & 0.9287 & 0.2633 & \textbf{0.3795} & 0.1087 & 0.5020 & 0.9323 & 0.1087 & \textbf{0.1419} & 0.0360 & 0.5000 & 0.0013 & 0.0360 & \textbf{0.0025} \\
\midrule

\multirow{4}{*}{\rotatebox[origin=c]{90}{Lung Cancer}} &
\multirow{4}{*}{ } &
\multirow{4}{*}{\rotatebox[origin=c]{90}{TRTR}} &
None & 
0.3000 & 0.3333 & 0.2600 & 0.3000 & 0.2750 & 0.2000 & 0.2222 & 0.2000 & 0.2000 & 0.2000 & 0.5000 & 0.5278 & 0.4583 & 0.5000 & \textbf{0.4714} \\
& & & Class Weight & 
0.3000 & 0.3333 & 0.2600 & 0.3000 & 0.2750 & 0.4000 & 0.4444 & 0.3000 & 0.4000 & 0.3429 & 0.5000 & 0.5278 & 0.4583 & 0.5000 & \textbf{0.4714} \\
& & & SMOTE & 
0.5000 & 0.5278 & 0.5000 & 0.5000 & \textbf{0.4762} & 0.6000 & 0.6389 & 0.6200 & 0.6000 & \textbf{0.5833} & 0.5000 & 0.5278 & 0.4583 & 0.5000 & \textbf{0.4714} \\
& & & Undersample & 
0.3000 & 0.3333 & 0.3750 & 0.3000 & 0.2591 & 0.5000 & 0.5556 & 0.3450 & 0.5000 & 0.4071 & 0.5000 & 0.5278 & 0.4633 & 0.5000 & 0.4593 \\
\midrule

\multirow{8}{*}{\rotatebox[origin=c]{90}{Lung Can.}} &
\multirow{8}{*}{\rotatebox[origin=c]{90}{ALM}} &
\multirow{4}{*}{\rotatebox[origin=c]{90}{TSTR}} &
None & 
0.6000 & 0.6389 & 0.5800 & 0.6000 & 0.5583 & 0.8000 & 0.8056 & 0.8250 & 0.8000 & \textbf{0.7971} & 0.7000 & 0.6944 & 0.7400 & 0.7000 & \textbf{0.7067} \\
& & & Class Weight & 
0.7000 & 0.7222 & 0.7467 & 0.7000 & \textbf{0.6936} & 0.6000 & 0.5833 & 0.6500 & 0.6000 & 0.6000 & 0.6000 & 0.6111 & 0.6000 & 0.6000 & 0.6000 \\
& & & SMOTE & 
0.6000 & 0.6389 & 0.5800 & 0.6000 & 0.5583 & 0.6000 & 0.5833 & 0.6500 & 0.6000 & 0.6000 & 0.6000 & 0.5833 & 0.6500 & 0.6000 & 0.6000 \\
& & & Undersample & 
0.4000 & 0.4444 & 0.3000 & 0.4000 & 0.3429 & 0.5000 & 0.5278 & 0.4833 & 0.5000 & 0.4857 & 0.4000 & 0.4444 & 0.3000 & 0.4000 & 0.3429 \\
\cmidrule(lr){3-3}
& & \multirow{4}{*}{\rotatebox[origin=c]{90}{TRTS}} &
None & 
0.5000 & 0.5000 & 0.5417 & 0.5000 & \textbf{0.5107} & 0.6250 & 0.5833 & 0.6250 & 0.6250 & 0.6250 & 0.5000 & 0.4167 & 0.3750 & 0.5000 & 0.4250 \\
& & & Class Weight & 
0.5000 & 0.5000 & 0.5417 & 0.5000 & \textbf{0.5107} & 0.7500 & 0.6667 & 0.8333 & 0.7500 & \textbf{0.7333} & 0.3750 & 0.3333 & 0.2833 & 0.3750 & 0.3222 \\
& & & SMOTE & 
0.2500 & 0.3333 & 0.1875 & 0.2500 & 0.2083 & 0.6250 & 0.5833 & 0.6250 & 0.6250 & 0.6250 & 0.6250 & 0.5833 & 0.6750 & 0.6250 & \textbf{0.6250} \\
& & & Undersample & 
0.3750 & 0.3333 & 0.5500 & 0.3750 & 0.4048 & 0.5000 & 0.5000 & 0.6667 & 0.5000 & 0.5333 & 0.5000 & 0.5000 & 0.6875 & 0.5000 & 0.5417 \\
\midrule

\multirow{8}{*}{\rotatebox[origin=c]{90}{Lung Cancer}} &
\multirow{8}{*}{\rotatebox[origin=c]{90}{RTF}} &
\multirow{4}{*}{\rotatebox[origin=c]{90}{TSTR}} &
None & 
0.2000 & 0.1944 & 0.4000 & 0.2000 & 0.2500 & 0.5000 & 0.5278 & 0.4633 & 0.5000 & 0.4593 & 0.5000 & 0.5000 & 0.5100 & 0.5000 & \textbf{0.4978} \\
& & & Class Weight & 
0.2000 & 0.1944 & 0.3800 & 0.2000 & 0.2389 & 0.5000 & 0.5278 & 0.4633 & 0.5000 & 0.4593 & 0.4000 & 0.4167 & 0.4083 & 0.4000 & 0.4000 \\
& & & SMOTE & 
0.5000 & 0.5000 & 0.6100 & 0.5000 & \textbf{0.4992} & 0.5000 & 0.5278 & 0.4583 & 0.5000 & \textbf{0.4714} & 0.4000 & 0.4167 & 0.4000 & 0.4000 & 0.4000 \\
& & & Undersample & 
0.2000 & 0.1944 & 0.1800 & 0.2000 & 0.1889 & 0.5000 & 0.5278 & 0.4583 & 0.5000 & \textbf{0.4714} & 0.4000 & 0.4167 & 0.4083 & 0.4000 & 0.4000 \\
\cmidrule(lr){3-3}
& & \multirow{4}{*}{\rotatebox[origin=c]{90}{TRTS}} &
None & 
0.3750 & 0.4444 & 0.3125 & 0.3750 & 0.3167 & 0.6250 & 0.6667 & 0.6250 & 0.6250 & 0.6143 & 0.7500 & 0.7778 & 0.7500 & 0.7500 & \textbf{0.7500} \\
& & & Class Weight & 
0.3750 & 0.4444 & 0.3125 & 0.3750 & 0.3167 & 0.6250 & 0.6667 & 0.6250 & 0.6250 & 0.6143 & 0.7500 & 0.7778 & 0.7500 & 0.7500 & \textbf{0.7500} \\
& & & SMOTE & 
0.5000 & 0.5556 & 0.3750 & 0.5000 & \textbf{0.4167} & 0.7500 & 0.7778 & 0.8229 & 0.7500 & 0.7089 & 0.7500 & 0.7778 & 0.7500 & 0.7500 & \textbf{0.7500} \\
& & & Undersample & 
0.3750 & 0.4444 & 0.4464 & 0.3750 & 0.2986 & 0.8750 & 0.8889 & 0.9062 & 0.8750 & \textbf{0.8714} & 0.6250 & 0.6667 & 0.6042 & 0.6250 & 0.6000 \\
\midrule

\multirow{4}{*}{\rotatebox[origin=c]{90}{Thyroid}} &
\multirow{4}{*}{ } &
\multirow{4}{*}{\rotatebox[origin=c]{90}{TRTR}} &
None & 
0.9832 & 0.9435 & 0.9833 & 0.9832 & \textbf{0.9833} & 0.9629 & 0.7902 & 0.9610 & 0.9629 & \textbf{0.9593} & 0.9461 & 0.7021 & 0.9399 & 0.9461 & \textbf{0.9383} \\
& & & Class Weight & 
0.9779 & 0.9406 & 0.9789 & 0.9779 & 0.9783 & 0.9532 & 0.7217 & 0.9505 & 0.9532 & 0.9460 & 0.8392 & 0.8392 & 0.9319 & 0.8392 & 0.8706 \\
& & & SMOTE & 
0.9638 & 0.9277 & 0.9689 & 0.9638 & 0.9655 & 0.9558 & 0.8760 & 0.9590 & 0.9558 & 0.9571 & 0.8595 & 0.8186 & 0.9282 & 0.8595 & 0.8838 \\
& & & Undersample & 
0.9170 & 0.9392 & 0.9571 & 0.9170 & 0.9291 & 0.8648 & 0.9163 & 0.9488 & 0.8648 & 0.8908 & 0.8189 & 0.8123 & 0.9260 & 0.8189 & 0.8557 \\
\midrule

\multirow{8}{*}{\rotatebox[origin=c]{90}{Thyroid}} &
\multirow{8}{*}{\rotatebox[origin=c]{90}{ALM}} &
\multirow{4}{*}{\rotatebox[origin=c]{90}{TSTR}} &
None & 
0.8507 & 0.7453 & 0.9126 & 0.8507 & 0.8746 & 0.9108 & 0.5249 & 0.8733 & 0.9108 & \textbf{0.8880} & 0.9152 & 0.4957 & 0.8516 & 0.9152 & 0.8823 \\
& & & Class Weight & 
0.9223 & 0.4995 & 0.8521 & 0.9223 & \textbf{0.8858} & 0.9231 & 0.5000 & 0.8522 & 0.9231 & 0.8863 & 0.9055 & 0.5273 & 0.8717 & 0.9055 & 0.8859 \\
& & & SMOTE & 
0.8269 & 0.5796 & 0.8764 & 0.8269 & 0.8491 & 0.9231 & 0.5000 & 0.8522 & 0.9231 & 0.8863 & 0.9231 & 0.5000 & 0.8522 & 0.9231 & \textbf{0.8863} \\
& & & Undersample & 
0.8375 & 0.8540 & 0.9351 & 0.8375 & 0.8698 & 0.8569 & 0.8013 & 0.9245 & 0.8569 & 0.8813 & 0.7915 & 0.7027 & 0.9022 & 0.7915 & 0.8334 \\
\cmidrule(lr){3-3}
& & \multirow{4}{*}{\rotatebox[origin=c]{90}{TRTS}} &
None & 
0.9536 & 0.8985 & 0.9664 & 0.9536 & 0.9581 & 0.9459 & 0.5000 & 0.8948 & 0.9459 & 0.9196 & 0.1336 & 0.5420 & 0.9491 & 0.1336 & 0.1526 \\
& & & Class Weight & 
0.9614 & 0.8834 & 0.9676 & 0.9614 & \textbf{0.9638} & 0.9459 & 0.5000 & 0.8948 & 0.9459 & 0.9196 & 0.0949 & 0.5216 & 0.9490 & 0.0949 & 0.0841 \\
& & & SMOTE & 
0.9183 & 0.4854 & 0.8933 & 0.9183 & 0.9056 & 0.9503 & 0.6082 & 0.9385 & 0.9503 & \textbf{0.9393} & 0.1965 & 0.5753 & 0.9493 & 0.1965 & \textbf{0.2539} \\
& & & Undersample & 
0.8510 & 0.8924 & 0.9560 & 0.8510 & 0.8873 & 0.9404 & 0.6222 & 0.9297 & 0.9404 & 0.9340 & 0.1777 & 0.5653 & 0.9493 & 0.1777 & 0.2250 \\
\midrule

\multirow{8}{*}{\rotatebox[origin=c]{90}{Thyroid}} &
\multirow{8}{*}{\rotatebox[origin=c]{90}{RTF}} &
\multirow{4}{*}{\rotatebox[origin=c]{90}{TSTR}} &
None & 
0.9028 & 0.6523 & 0.9018 & 0.9028 & \textbf{0.9023} & 0.9231 & 0.5000 & 0.8522 & 0.9231 & \textbf{0.8863} & 0.8127 & 0.6773 & 0.8969 & 0.8127 & 0.8463 \\
& & & Class Weight & 
0.9108 & 0.6039 & 0.8953 & 0.9108 & 0.9018 & 0.9231 & 0.5000 & 0.8522 & 0.9231 & \textbf{0.8863} & 0.5848 & 0.7277 & 0.9203 & 0.5848 & 0.6774 \\
& & & SMOTE & 
0.8719 & 0.5302 & 0.8670 & 0.8719 & 0.8694 & 0.9231 & 0.5000 & 0.8522 & 0.9231 & \textbf{0.8863} & 0.9231 & 0.5000 & 0.8522 & 0.9231 & \textbf{0.8863} \\
& & & Undersample & 
0.8348 & 0.8947 & 0.9440 & 0.8348 & 0.8691 & 0.6670 & 0.8091 & 0.9346 & 0.6670 & 0.7443 & 0.2235 & 0.5794 & 0.9301 & 0.2235 & 0.2658 \\
\cmidrule(lr){3-3}
& & \multirow{4}{*}{\rotatebox[origin=c]{90}{TRTS}} &
None & 
0.9636 & 0.9126 & 0.9699 & 0.9636 & 0.9658 & 0.9404 & 0.5091 & 0.9440 & 0.9404 & 0.9126 & 0.1280 & 0.5358 & 0.9432 & 0.1280 & 0.1331 \\
& & & Class Weight & 
0.9669 & 0.8973 & 0.9702 & 0.9669 & \textbf{0.9682} & 0.9393 & 0.5000 & 0.8823 & 0.9393 & 0.9099 & 0.0993 & 0.5206 & 0.9431 & 0.0993 & 0.0814 \\
& & & SMOTE & 
0.9139 & 0.4865 & 0.8808 & 0.9139 & 0.8970 & 0.9404 & 0.5091 & 0.9440 & 0.9404 & 0.9126 & 0.1600 & 0.5529 & 0.9434 & 0.1600 & 0.1873 \\
& & & Undersample & 
0.8653 & 0.9198 & 0.9567 & 0.8653 & 0.8954 & 0.9294 & 0.5797 & 0.9120 & 0.9294 & \textbf{0.9190} & 0.1667 & 0.5564 & 0.9434 & 0.1667 & \textbf{0.1982} \\
\midrule

\multirow{4}{*}{\rotatebox[origin=c]{90}{Liver Dis.}} &
\multirow{4}{*}{ } &
\multirow{4}{*}{\rotatebox[origin=c]{90}{TRTR}} &
None & 
0.6442 & 0.6492 & 0.6580 & 0.6442 & 0.6463 & 0.7692 & 0.7606 & 0.7682 & 0.7692 & \textbf{0.7684} & 0.7212 & 0.7068 & 0.7189 & 0.7212 & \textbf{0.7183} \\
& & & Class Weight & 
0.5673 & 0.5735 & 0.5845 & 0.5673 & 0.5695 & 0.7596 & 0.7523 & 0.7590 & 0.7596 & 0.7592 & 0.6635 & 0.6720 & 0.6818 & 0.6635 & 0.6652 \\
& & & SMOTE & 
0.5769 & 0.5758 & 0.5855 & 0.5769 & 0.5793 & 0.7404 & 0.7386 & 0.7435 & 0.7404 & 0.7413 & 0.6635 & 0.6720 & 0.6818 & 0.6635 & 0.6652 \\
& & & Undersample & 
0.6538 & 0.6818 & 0.7172 & 0.6538 & \textbf{0.6477} & 0.7019 & 0.7023 & 0.7082 & 0.7019 & 0.7035 & 0.6635 & 0.6720 & 0.6818 & 0.6635 & 0.6652 \\
\midrule

\multirow{8}{*}{\rotatebox[origin=c]{90}{Liver Dis.}} &
\multirow{8}{*}{\rotatebox[origin=c]{90}{ALM}} &
\multirow{4}{*}{\rotatebox[origin=c]{90}{TSTR}} &
None & 
0.4615 & 0.4515 & 0.4647 & 0.4615 & 0.4630 & 0.5288 & 0.5220 & 0.5332 & 0.5288 & 0.5306 & 0.5577 & 0.5136 & 0.5300 & 0.5577 & 0.5182 \\
& & & Class Weight & 
0.5481 & 0.5417 & 0.5523 & 0.5481 & \textbf{0.5497} & 0.5192 & 0.4924 & 0.5039 & 0.5192 & 0.5069 & 0.5577 & 0.5591 & 0.5695 & 0.5577 & 0.5603 \\
& & & SMOTE & 
0.5385 & 0.5394 & 0.5503 & 0.5385 & 0.5412 & 0.4904 & 0.4856 & 0.4979 & 0.4904 & 0.4930 & 0.5577 & 0.5591 & 0.5695 & 0.5577 & 0.5603 \\
& & & Undersample & 
0.5192 & 0.5197 & 0.5311 & 0.5192 & 0.5221 & 0.5385 & 0.5455 & 0.5570 & 0.5385 & \textbf{0.5405} & 0.5673 & 0.5765 & 0.5884 & 0.5673 & \textbf{0.5689} \\
\cmidrule(lr){3-3}
& & \multirow{4}{*}{\rotatebox[origin=c]{90}{TRTS}} &
None & 
0.4940 & 0.4890 & 0.5016 & 0.4940 & 0.4967 & 0.5542 & 0.5372 & 0.5492 & 0.5542 & 0.5510 & 0.4940 & 0.4658 & 0.4763 & 0.4940 & \textbf{0.4808} \\
& & & Class Weight & 
0.4699 & 0.4798 & 0.4918 & 0.4699 & 0.4708 & 0.5663 & 0.5515 & 0.5630 & 0.5663 & 0.5643 & 0.3976 & 0.3902 & 0.4058 & 0.3976 & 0.4008 \\
& & & SMOTE & 
0.5060 & 0.4994 & 0.5117 & 0.5060 & \textbf{0.5082} & 0.5542 & 0.5411 & 0.5525 & 0.5542 & 0.5533 & 0.4096 & 0.4045 & 0.4195 & 0.4096 & 0.4131 \\
& & & Undersample & 
0.4819 & 0.4824 & 0.4951 & 0.4819 & 0.4851 & 0.5783 & 0.5735 & 0.5835 & 0.5783 & \textbf{0.5802} & 0.4096 & 0.4006 & 0.4159 & 0.4096 & 0.4122 \\
\midrule

\multirow{8}{*}{\rotatebox[origin=c]{90}{Liver Dis.}} &
\multirow{8}{*}{\rotatebox[origin=c]{90}{RTF}} &
\multirow{4}{*}{\rotatebox[origin=c]{90}{TSTR}} &
None & 
0.5385 & 0.5394 & 0.5503 & 0.5385 & \textbf{0.5412} & 0.5577 & 0.5439 & 0.5551 & 0.5577 & 0.5562 & 0.6154 & 0.5455 & 0.7692 & 0.6154 & 0.5032 \\
& & & Class Weight & 
0.4904 & 0.4856 & 0.4979 & 0.4904 & 0.4930 & 0.5673 & 0.5553 & 0.5660 & 0.5673 & \textbf{0.5666} & 0.5962 & 0.6106 & 0.6251 & 0.5962 & 0.5962 \\
& & & SMOTE & 
0.4519 & 0.4553 & 0.4677 & 0.4519 & 0.4547 & 0.5192 & 0.5106 & 0.5221 & 0.5192 & 0.5205 & 0.5962 & 0.5955 & 0.6046 & 0.5962 & 0.5984 \\
& & & Undersample & 
0.5096 & 0.5205 & 0.5326 & 0.5096 & 0.5103 & 0.5385 & 0.5485 & 0.5607 & 0.5385 & 0.5397 & 0.6731 & 0.6773 & 0.6849 & 0.6731 & \textbf{0.6750} \\
\cmidrule(lr){3-3}
& & \multirow{4}{*}{\rotatebox[origin=c]{90}{TRTS}} &
None & 
0.5422 & 0.5307 & 0.5392 & 0.5422 & 0.5404 & 0.5783 & 0.5529 & 0.5673 & 0.5783 & 0.5624 & 0.6265 & 0.6085 & 0.6213 & 0.6265 & \textbf{0.6193} \\
& & & Class Weight & 
0.5301 & 0.5168 & 0.5256 & 0.5301 & 0.5272 & 0.5904 & 0.5700 & 0.5824 & 0.5904 & \textbf{0.5808} & 0.6024 & 0.5937 & 0.6012 & 0.6024 & 0.6017 \\
& & & SMOTE & 
0.5181 & 0.5127 & 0.5212 & 0.5181 & 0.5193 & 0.5904 & 0.5700 & 0.5824 & 0.5904 & \textbf{0.5808} & 0.5904 & 0.5863 & 0.5932 & 0.5904 & 0.5914 \\
& & & Undersample & 
0.5542 & 0.5609 & 0.5697 & 0.5542 & \textbf{0.5554} & 0.5783 & 0.5691 & 0.5770 & 0.5783 & 0.5776 & 0.5904 & 0.5863 & 0.5932 & 0.5904 & 0.5914 \\
\midrule

\multirow{4}{*}{\rotatebox[origin=c]{90}{Heart Fail.}} &
\multirow{4}{*}{ } &
\multirow{4}{*}{\rotatebox[origin=c]{90}{TRTR}} &
None & 
0.7889 & 0.7448 & 0.7844 & 0.7889 & \textbf{0.7857} & 0.8111 & 0.7612 & 0.8065 & 0.8111 & 0.8059 & 0.8222 & 0.7694 & 0.8187 & 0.8222 & \textbf{0.8161} \\
& & & Class Weight & 
0.7556 & 0.7021 & 0.7484 & 0.7556 & 0.7504 & 0.8444 & 0.8038 & 0.8417 & 0.8444 & 0.8412 & 0.7889 & 0.7538 & 0.7871 & 0.7889 & 0.7879 \\
& & & SMOTE & 
0.7778 & 0.7456 & 0.7778 & 0.7778 & 0.7778 & 0.8556 & 0.8392 & 0.8570 & 0.8556 & \textbf{0.8562} & 0.8000 & 0.7801 & 0.8042 & 0.8000 & 0.8017 \\
& & & Undersample & 
0.7222 & 0.7408 & 0.7695 & 0.7222 & 0.7311 & 0.7889 & 0.8081 & 0.8249 & 0.7889 & 0.7953 & 0.7778 & 0.7637 & 0.7877 & 0.7778 & 0.7812 \\
\midrule

\multirow{8}{*}{\rotatebox[origin=c]{90}{Heart Fail.}} &
\multirow{8}{*}{\rotatebox[origin=c]{90}{ALM}} &
\multirow{4}{*}{\rotatebox[origin=c]{90}{TSTR}} &
None & 
0.6667 & 0.6003 & 0.6552 & 0.6667 & \textbf{0.6596} & 0.7333 & 0.6133 & 0.7320 & 0.7333 & 0.6905 & 0.7556 & 0.6207 & 0.8203 & 0.7556 & 0.6995 \\
& & & Class Weight & 
0.6778 & 0.5633 & 0.6427 & 0.6778 & 0.6392 & 0.7000 & 0.5435 & 0.7145 & 0.7000 & 0.6119 & 0.6556 & 0.6464 & 0.6870 & 0.6556 & 0.6650 \\
& & & SMOTE & 
0.6444 & 0.6382 & 0.6803 & 0.6444 & 0.6547 & 0.7222 & 0.6504 & 0.7087 & 0.7222 & \textbf{0.7107} & 0.7000 & 0.6792 & 0.7152 & 0.7000 & \textbf{0.7055} \\
& & & Undersample & 
0.5000 & 0.5226 & 0.5840 & 0.5000 & 0.5145 & 0.6222 & 0.6399 & 0.6854 & 0.6222 & 0.6346 & 0.5889 & 0.5791 & 0.6304 & 0.5889 & 0.6013 \\
\cmidrule(lr){3-3}
& & \multirow{4}{*}{\rotatebox[origin=c]{90}{TRTS}} &
None & 
0.5694 & 0.5245 & 0.5957 & 0.5694 & 0.5801 & 0.6667 & 0.5818 & 0.6502 & 0.6667 & \textbf{0.6564} & 0.5833 & 0.4964 & 0.5724 & 0.5833 & 0.5775 \\
& & & Class Weight & 
0.5694 & 0.4991 & 0.5749 & 0.5694 & 0.5721 & 0.6389 & 0.5618 & 0.6297 & 0.6389 & 0.6338 & 0.5833 & 0.5600 & 0.6246 & 0.5833 & 0.5973 \\
& & & SMOTE & 
0.5972 & 0.5445 & 0.6124 & 0.5972 & \textbf{0.6038} & 0.6111 & 0.5418 & 0.6111 & 0.6111 & 0.6111 & 0.6111 & 0.5545 & 0.6209 & 0.6111 & \textbf{0.6156} \\
& & & Undersample & 
0.5556 & 0.5655 & 0.6312 & 0.5556 & 0.5730 & 0.5972 & 0.5573 & 0.6224 & 0.5972 & 0.6072 & 0.5556 & 0.5400 & 0.6085 & 0.5556 & 0.5720 \\
\midrule

\multirow{8}{*}{\rotatebox[origin=c]{90}{Heart Fail.}} &
\multirow{8}{*}{\rotatebox[origin=c]{90}{RTF}} &
\multirow{4}{*}{\rotatebox[origin=c]{90}{TSTR}} &
None & 
0.5444 & 0.5011 & 0.5642 & 0.5444 & 0.5528 & 0.6556 & 0.5017 & 0.5673 & 0.6556 & 0.5697 & 0.6778 & 0.5090 & 0.6232 & 0.6778 & 0.5666 \\
& & & Class Weight & 
0.4667 & 0.3985 & 0.4759 & 0.4667 & 0.4711 & 0.6667 & 0.5008 & 0.5670 & 0.6667 & 0.5605 & 0.5778 & 0.5890 & 0.6410 & 0.5778 & 0.5916 \\
& & & SMOTE & 
0.5111 & 0.4856 & 0.5510 & 0.5111 & 0.5252 & 0.6000 & 0.4878 & 0.5497 & 0.6000 & 0.5657 & 0.5333 & 0.5291 & 0.5883 & 0.5333 & 0.5485 \\
& & & Undersample & 
0.6444 & 0.6653 & 0.7077 & 0.6444 & \textbf{0.6560} & 0.5667 & 0.5899 & 0.6439 & 0.5667 & \textbf{0.5802} & 0.6556 & 0.6645 & 0.7041 & 0.6556 & \textbf{0.6665} \\
\cmidrule(lr){3-3}
& & \multirow{4}{*}{\rotatebox[origin=c]{90}{TRTS}} &
None & 
0.5417 & 0.5135 & 0.6091 & 0.5417 & 0.5640 & 0.6667 & 0.5692 & 0.6564 & 0.6667 & 0.6610 & 0.6528 & 0.5288 & 0.6248 & 0.6528 & 0.6359 \\
& & & Class Weight & 
0.5972 & 0.5212 & 0.6152 & 0.5972 & \textbf{0.6053} & 0.6806 & 0.5942 & 0.6757 & 0.6806 & \textbf{0.6780} & 0.6250 & 0.5558 & 0.6420 & 0.6250 & 0.6325 \\
& & & SMOTE & 
0.5833 & 0.5885 & 0.6682 & 0.5833 & 0.6049 & 0.6667 & 0.6000 & 0.6770 & 0.6667 & 0.6713 & 0.6528 & 0.5750 & 0.6581 & 0.6528 & 0.6553 \\
& & & Undersample & 
0.4861 & 0.5519 & 0.6481 & 0.4861 & 0.5036 & 0.5694 & 0.5481 & 0.6357 & 0.5694 & 0.5904 & 0.6528 & 0.6058 & 0.6797 & 0.6528 & \textbf{0.6633} \\
\midrule

\multirow{4}{*}{\rotatebox[origin=c]{90}{PIR Vis.}} &
\multirow{4}{*}{ } &
\multirow{4}{*}{\rotatebox[origin=c]{90}{TRTR}} &
None & 
0.9874 & 0.9798 & 0.9874 & 0.9874 & \textbf{0.9874} & 0.9917 & 0.9816 & 0.9917 & 0.9917 & 0.9917 & 0.9599 & 0.9030 & 0.9603 & 0.9599 & \textbf{0.9598} \\
& & & Class Weight & 
0.9852 & 0.9766 & 0.9853 & 0.9852 & 0.9852 & 0.9930 & 0.9844 & 0.9930 & 0.9930 & \textbf{0.9930} & 0.9334 & 0.9341 & 0.9509 & 0.9334 & 0.9381 \\
& & & SMOTE & 
0.9813 & 0.9796 & 0.9823 & 0.9813 & 0.9816 & 0.9900 & 0.9901 & 0.9903 & 0.9900 & 0.9901 & 0.9395 & 0.9319 & 0.9535 & 0.9395 & 0.9433 \\
& & & Undersample & 
0.9787 & 0.9774 & 0.9801 & 0.9787 & 0.9791 & 0.9804 & 0.9862 & 0.9824 & 0.9804 & 0.9809 & 0.9290 & 0.9323 & 0.9490 & 0.9290 & 0.9344 \\
\midrule

\multirow{8}{*}{\rotatebox[origin=c]{90}{PIR Vis.}} &
\multirow{8}{*}{\rotatebox[origin=c]{90}{ALM}} &
\multirow{4}{*}{\rotatebox[origin=c]{90}{TSTR}} &
None & 
0.9839 & 0.9784 & 0.9843 & 0.9839 & 0.9840 & 0.9909 & 0.9836 & 0.9909 & 0.9909 & 0.9908 & 0.9730 & 0.9232 & 0.9740 & 0.9730 & \textbf{0.9728} \\
& & & Class Weight & 
0.9869 & 0.9831 & 0.9872 & 0.9869 & \textbf{0.9870} & 0.9917 & 0.9874 & 0.9918 & 0.9917 & \textbf{0.9917} & 0.9395 & 0.9328 & 0.9570 & 0.9395 & 0.9437 \\
& & & SMOTE & 
0.9856 & 0.9895 & 0.9868 & 0.9856 & 0.9859 & 0.9882 & 0.9871 & 0.9886 & 0.9882 & 0.9883 & 0.9578 & 0.9402 & 0.9660 & 0.9578 & 0.9595 \\
& & & Undersample & 
0.9739 & 0.9847 & 0.9780 & 0.9739 & 0.9749 & 0.9795 & 0.9847 & 0.9816 & 0.9795 & 0.9801 & 0.9425 & 0.9376 & 0.9582 & 0.9425 & 0.9462 \\
\cmidrule(lr){3-3}
& & \multirow{4}{*}{\rotatebox[origin=c]{90}{TRTS}} &
None & 
0.9771 & 0.9622 & 0.9769 & 0.9771 & 0.9769 & 0.9886 & 0.9831 & 0.9885 & 0.9886 & \textbf{0.9885} & 0.9134 & 0.8235 & 0.9163 & 0.9134 & \textbf{0.9126} \\
& & & Class Weight & 
0.9739 & 0.9539 & 0.9735 & 0.9739 & 0.9734 & 0.9820 & 0.9664 & 0.9819 & 0.9820 & 0.9818 & 0.8715 & 0.8349 & 0.8997 & 0.8715 & 0.8784 \\
& & & SMOTE & 
0.9733 & 0.9615 & 0.9731 & 0.9733 & 0.9732 & 0.9842 & 0.9782 & 0.9842 & 0.9842 & 0.9842 & 0.8878 & 0.8446 & 0.9103 & 0.8878 & 0.8928 \\
& & & Undersample & 
0.9782 & 0.9725 & 0.9784 & 0.9782 & \textbf{0.9783} & 0.9788 & 0.9855 & 0.9805 & 0.9788 & 0.9792 & 0.8715 & 0.8385 & 0.9081 & 0.8715 & 0.8798 \\
\midrule

\multirow{8}{*}{\rotatebox[origin=c]{90}{PIR Vis.}} &
\multirow{8}{*}{\rotatebox[origin=c]{90}{RTF}} &
\multirow{4}{*}{\rotatebox[origin=c]{90}{TSTR}} &
None & 
0.9826 & 0.9675 & 0.9825 & 0.9826 & \textbf{0.9825} & 0.9891 & 0.9748 & 0.9891 & 0.9891 & 0.9890 & 0.9608 & 0.8713 & 0.9611 & 0.9608 & 0.9596 \\
& & & Class Weight & 
0.9813 & 0.9646 & 0.9811 & 0.9813 & 0.9812 & 0.9900 & 0.9786 & 0.9899 & 0.9900 & 0.9899 & 0.9499 & 0.9047 & 0.9581 & 0.9499 & 0.9512 \\
& & & SMOTE & 
0.9813 & 0.9692 & 0.9814 & 0.9813 & 0.9813 & 0.9904 & 0.9880 & 0.9906 & 0.9904 & \textbf{0.9905} & 0.9564 & 0.8974 & 0.9610 & 0.9564 & 0.9567 \\
& & & Undersample & 
0.9708 & 0.9800 & 0.9753 & 0.9708 & 0.9720 & 0.9834 & 0.9852 & 0.9846 & 0.9834 & 0.9838 & 0.9591 & 0.9079 & 0.9648 & 0.9591 & \textbf{0.9597} \\
\cmidrule(lr){3-3}
& & \multirow{4}{*}{\rotatebox[origin=c]{90}{TRTS}} &
None & 
0.9820 & 0.9630 & 0.9818 & 0.9820 & 0.9819 & 0.9902 & 0.9841 & 0.9902 & 0.9902 & \textbf{0.9902} & 0.9532 & 0.8708 & 0.9550 & 0.9532 & \textbf{0.9533} \\
& & & Class Weight & 
0.9793 & 0.9633 & 0.9793 & 0.9793 & 0.9793 & 0.9864 & 0.9692 & 0.9862 & 0.9864 & 0.9862 & 0.9134 & 0.8825 & 0.9376 & 0.9134 & 0.9203 \\
& & & SMOTE & 
0.9744 & 0.9688 & 0.9759 & 0.9744 & 0.9749 & 0.9886 & 0.9820 & 0.9887 & 0.9886 & 0.9886 & 0.9232 & 0.8805 & 0.9414 & 0.9232 & 0.9280 \\
& & & Undersample & 
0.9831 & 0.9783 & 0.9837 & 0.9831 & \textbf{0.9833} & 0.9853 & 0.9866 & 0.9862 & 0.9853 & 0.9856 & 0.9112 & 0.8683 & 0.9389 & 0.9112 & 0.9186 \\

\end{longtable}
\endgroup
\clearpage
\twocolumn



\section{Reproducibility and Computational Setup}
\label{sec:reproducibility}

To ensure the full reproducibility of our empirical findings, all experiments, model training, and evaluation pipelines were executed in a strictly controlled and documented computational environment. 

\paragraph{Hardware Infrastructure.}
All generative model training, both the Augmented Lagrangian Method constraint optimizations and the baseline autoregressive/GAN training, and downstream adversarial evaluations were accelerated using NVIDIA GPU A10 hardware (24GB VRAM), an AMD EPYC 7742 64-Core Processor, and 256 GB of system RAM. 

\paragraph{Software Dependencies.}
The PSyGenTAB framework and all baseline models were implemented in Python 3.10. The core deep learning architectures and ALM optimization loops were built using \texttt{PyTorch} (v2.1.0) with CUDA 12.2 support. Tabular data preprocessing, imputation, and transformation were handled via \texttt{pandas} (v2.1.1) and \texttt{scikit-learn} (v1.3.1). 

For the baseline generative architectures, we utilized the official open-source implementations of RealTabFormer (via the \texttt{transformers} library v4.34.0) and CTAB-GAN+. 

\paragraph{Evaluation Libraries.}
The multi-dimensional evaluation metrics were computed using standardized, community-vetted libraries to ensure fairness and consistency:
\begin{itemize}
    \item \textbf{Statistical Quality and Fidelity:} Computed using the \texttt{SDMetrics} (v0.12.0) library for open-source structural checks (e.g., KS Test, TVD) and the \texttt{mostlyai-qa} package for commercial-grade multivariate accuracy scoring.
    \item \textbf{Downstream ML Utility (TSTR/TRTS):} Implemented using \texttt{scikit-learn}. Decision Trees, Random Forests, and Logistic Regression models were instantiated with default hyperparameters to evaluate the inherent signal of the data rather than model-tuning prowess. Imbalance strategies utilized the \texttt{imbalanced-learn} (v0.11.0) library (specifically for SMOTE and RandomUnderSampler).
    \item \textbf{Privacy and Adversarial Risk:} Distance to Closest Record (DCR), Exact Match Ratios, and Membership Inference Attacks (MIA) were evaluated using a combination of custom vectorized distance calculations (for computational efficiency on large sets like Diabetes) and the \texttt{Anonymeter} privacy evaluation suite.
\end{itemize}

\paragraph{Code and Data Availability.}
To facilitate community adoption and verify our claims, the complete source code for PSyGenTAB—including the ALM objective functions, the FAITH validation scripts, and the hyperparameter configurations for all datasets were made publicly available. Repository:\url{https://github.com/ArshiaIlaty/PsyGenTAB}

\end{document}